\documentclass[sigconf,screen,nonacm]{acmart}

\usepackage{pdflscape}
\usepackage{enumitem}
\usepackage{array}
\usepackage{subcaption}
\usepackage{tabularx}
\usepackage{makecell}
\usepackage{threeparttable}
\usepackage{listings}
\usepackage{tikz}
\usepackage{placeins}
\usepackage{setspace}

\AtBeginDocument{%
  }

\pgfsetarrows{latex-latex}
\tikzset{%
  base/.style = {
    inner sep=5pt,
    text centered,
    thin,
    font=\rmfamily
  },
  round/.style = {
    base,
    rectangle,
    rounded corners=1ex,
    draw=black,
    fill=gray!20,
    minimum height=0.35in
  }
}

\setcopyright{none}
\acmDOI{}
\acmYear{2027}

\acmConference[CSLAW '27]
  {6th ACM Symposium on Computer Science and Law}
  {March 23--25, 2027}
  {Chicago, Illinois, USA}

\begin{document}

\title{DECODEM: Data Extraction from Corporate Organizational Documents via Enhanced Methods}

\author{Jens Frankenreiter}
\email{fjens@wustl.edu}
\orcid{0000-0003-0541-8521}
\affiliation{%
  \institution{Washington University in St. Louis School of Law}
  \city{St. Louis}
  \state{Missouri}
  \country{USA}
}

\begin{abstract}
Much empirical legal research depends on translating unstructured text into structured variables. In corporate governance research as elsewhere, this translation has traditionally been performed by humans. Manual coding is time-consuming and costly. In corporate governance research, standardized commercial datasets offer an alternative, but they cover a limited set of firms and governance variables.

This paper introduces the DECODEM Benchmarks and uses them to evaluate whether large language models can extract corporate governance variables from charters and bylaws. The benchmarks combine a sample of corporate charters and bylaws of publicly traded U.S. corporations obtained from the SEC’s EDGAR database with hand-coded labels that capture whether a document includes a specific governance arrangement. The paper uses these benchmarks to evaluate a range of extraction approaches and models. The approaches vary along several dimensions, including the amount of guidance included in the prompts and the extent to which the model is provided with the full document or with relevant excerpts only.

The results show that current models can recover many governance variables with high accuracy across extraction approaches. At the same time, performance varies substantially across documents and variables. In particular, extraction is less accurate for charters than for bylaws, including when the comparison is limited to variables that appear in both benchmarks. The results also show that frontier models perform comparably on the tasks evaluated here and that more elaborate extraction pipelines do not consistently improve their performance. More elaborate extraction strategies matter more for efficiency-oriented models, where they can substantially narrow the performance gap relative to frontier models. 
\end{abstract}

\maketitle

\section{Introduction}

Much empirical legal research depends on translating unstructured text into structured variables. This is particularly true in law and finance, where decades of work have studied the real-world implications of corporate governance arrangements (e.g., \cite{gompers2003corporate_governance,bebchuk2009what_matters}). Traditionally, this translation has been performed by human coders, who read legal documents such as corporate charters and bylaws and extract the relevant information following specified coding rules \citep{hall2008systematic_content_analysis}. This process is costly and difficult to scale, and many researchers obtain their data from standardized commercial datasets instead. However, these datasets are limited in the firms and variables they cover, which imposes important constraints on empirical research\citep{frankenreiter2025other_delaware_effect,kastiel2022corporate_governance_gap}. Recent advances in natural language processing raise the question whether this step can be automated.

Extracting legally relevant information from legal texts such as corporate charters and bylaws is difficult. These documents are often lengthy, use specialized language, and have a complex structure \citep{Ariai_2025,premasiri2025legalie}. Even small differences in wording can have important legal consequences. The meaning of a provision can also depend on its context, for example on definitions that appear elsewhere in a document. Earlier NLP methods often struggled with these features. The advent of large language models has made automated extraction from such documents much more promising, and prior research has successfully used them to recover governance variables at scale \citep{frankenreiter2025other_delaware_effect,frankenreiter2026sticky_charters}. Yet there is still little systematic evidence on how reliably they can recover legally meaningful variables from corporate charters and bylaws.

This paper introduces two benchmark datasets (the DECODEM benchmarks) designed to evaluate the ability of large language models to extract governance features from these documents. The benchmarks combine a sample of corporate charters and bylaws of publicly traded U.S. corporations obtained from the SEC's EDGAR database with hand-coded labels that capture whether a document includes a specific governance arrangement. The labels cover a broad range of governance provisions that have played a central role in empirical research, including provisions governing board structure, shareholder meetings, director elections, and indemnification and exculpation.

The paper uses these benchmarks to evaluate a range of extraction approaches and models. The approaches vary along several dimensions, including the amount of guidance included in the prompts and the extent to which the model is provided with the full document or with relevant excerpts only. The analysis also varies the language models used for the task, with a focus on one frontier and one efficiency-oriented model from each of the three major providers of proprietary large language models: OpenAI, Anthropic, and Google. This setup not only makes it possible to evaluate the main question whether the automatic extraction of governance variables is feasible. It also allows me to examine whether results are robust across different prompts and choices about pipeline design, and whether the findings generalize across models.

The results show that current models can recover many governance variables with high accuracy across extraction approaches. Performance also does not vary much across frontier models. Efficiency-oriented models, by comparison, benefit more from more elaborate extraction approaches, which can help at least partly close a substantial gap between frontier and efficiency-oriented models that emerges for more straightforward approaches. At the same time, performance varies substantially across documents and variables. In particular, extraction is less accurate for charters than for bylaws, including when the comparison is limited to variables that appear in both benchmarks. A small number of provisions that are particularly difficult to extract from charters---such as shareholder meeting rights, written consent, and vacancy-filling authority---account for much of this difference.

Overall, the paper makes three contributions. First, it introduces the DECODEM benchmarks, a pair of benchmark datasets for evaluating information extraction from corporate charters and bylaws. Second, it provides a comparison of extraction architectures and language models on the classification tasks defined by the DECODEM benchmarks. Third, it documents systematic differences across document types, showing that the same governance variables are generally more difficult to extract from charters than from bylaws. A companion paper \citep{frankenreiter2026measuring} uses the same benchmarks to address a related set of questions, with a focus on comparing automated extraction against realistic human coding workflows.
\section{Text-to-Variable Translation in Empirical Corporate Governance Research}

A central step in many empirical legal research projects is transforming unstructured text into structured variables \citep{hall2008systematic_content_analysis,gentzkow2019text}. The rules and institutional arrangements researchers intend to study are often contained in textual sources such as contracts, statutes, regulatory filings, and court opinions. But the relevant information is rarely in a format that would directly lend itself to statistical analysis. It must first be translated into structured data, typically by identifying whether particular features appear in a document and mapping them to categorical or binary variables under predefined coding rules.

While this challenge arises in many settings, this paper studies it in the context of empirical corporate governance research. A large body of work in law and finance relies on variables capturing the governance arrangements of public corporations---director elections, shareholder rights, takeover defenses, and litigation-related provisions\citep{bebchuk2009what_matters,gompers2003corporate_governance}. Much of this information lives in corporate charters and bylaws, which define key elements of a firm's governance framework. To be used in empirical legal studies, this information must be translated into structured variables.

Traditionally, this translation has been performed by humans. Trained coders read documents and label the presence or absence of specific features\citep{hall2008systematic_content_analysis}. Manual coding is time-consuming and costly, so datasets tend to cover limited samples, and projects with uncertain payoffs may not be attempted at all. One partial solution is the creation of standardized datasets that make coded information available to many researchers working on a particular topic. These datasets can eliminate coding costs for secondary users, but at the price of a fixed menu of variables that may not match the needs of every potential user. In empirical corporate governance, standardized commercial datasets play a particularly important role\citep{bebchuk2009what_matters,gompers2003corporate_governance}, while hand collection has featured in fewer contributions, likely due to the high effort associated with manual coding\cite{frankenreiter2021cleaning_governance}. 

As a result, large parts of the empirical corporate governance literature focus on the firms covered in commercial datasets and the variables made available there\citep{kastiel2022corporate_governance_gap}. With only a small set of governance variables available, researchers focus on questions those variables can help answer. Many potentially important arrangements remain hard to study at scale because the relevant information has never been systematically coded. 

Against this background, a natural question is whether the translation from text to data, and in particular the extraction of corporate governance variables from charters and bylaws, can be automated. If so, researchers could extract previously unavailable variables from such documents at scale without the cost and effort associated with large-scale manual coding projects. Such data would make it possible to study arrangements that are missing from standard sources and expand coverage beyond the firms usually included in commercial datasets. Researchers could also use these data to audit existing datasets, whose accuracy has been questioned \citep{LarckerReissXiao2015,frankenreiter2021cleaning_governance}.

Until recently, the automated extraction of governance features faced major challenges. Legal documents are often lengthy, use specialized language, and have a complex structure. Relevant information may be spread across clauses or sections and may depend on small differences in phrasing \citep{beltagy2020longformerlongdocumenttransformer,NEURIPS2020_c8512d14}. This makes it difficult to rely on keyword- and rule-based extraction approaches. Earlier NLP methods worked well when the relevant information could be identified from relatively standardized wording \citep{nyarko2021stickiness_incomplete_contracts,rauterberg2017contracting_innovation}, but were less reliable when extraction required interpreting legal meaning or considering surrounding context. Many machine-learning-based NLP methods also require large labeled training sets, which are often unavailable.

Modern large language models offer an opportunity to overcome these challenges. They can account for context and for differences in meaning that are not captured by simple lexical patterns. They also need little or no task-specific labeled training data in many settings\citep{NEURIPS2020_1457c0d6}. In principle, this allows structured information to be extracted directly from complex documents. Recent work shows that this is not just a theoretical option: large language models can help extract governance features that are otherwise hard to observe systematically\citep{frankenreiter2026sticky_charters,frankenreiter2025other_delaware_effect}.

At the same time, despite the impressive progress large language models have made over the past years, high performance on this type of task is not a given. The responsible deployment of these methods requires the ability to measure extraction accuracy. While validation data can be assembled in the context of individual projects, their scope will usually be limited.

Benchmark datasets like the ones assembled here allow researchers to compare performance across models and extraction approaches using a common set of tasks. While such benchmarks cannot replace validation for a particular application, they provide a common basis for evaluating and comparing extraction methods. They can also help identify the types of provisions that remain difficult to extract, and they can help assess the sensitivity of extraction approaches to changes in prompt structure and pipeline design. Assembling them can therefore provide an important basis for understanding the strengths and limitations of automated coding.

Moreover, charters and bylaws provide a demanding test case for modern NLP. Benchmarks built from these documents therefore serve two purposes. They support empirical corporate governance research, but also provide a broader test case for information extraction from long and complex legal documents.
\section{Related Work}

This paper contributes to several strands of literature. First, a notable literature in natural language processing has developed benchmark datasets for a wide range of tasks, including legal tasks (e.g., \cite{guha2023legalbench,chalkidis-etal-2022-lexglue,hariri2026,afane2026benchmarking}). However, these benchmarks typically focus on tasks designed to assess a model's ability to perform forms of legal reasoning rather than on the extraction of structured variables from lengthy organizational documents. Many empirical legal research projects present a different challenge. They require researchers to process long, structurally complex texts and recover legally relevant information in a form suitable for quantitative analysis. The DECODEM Benchmarks are designed to evaluate model performance on precisely this type of task.

In terms of task design, the closest neighbor to DECODEM is Marotta-Wurgler and Stein’s privacy-policy corpus, which develops a hand-coded dataset and coding toolkit for translating lengthy legal documents into structured variables \citep{marottaWurglerStein2025PrivacyPolicyCorpus}. Other contract-related benchmarks are also relevant insofar as, like DECODEM, they focus on identifying or classifying provisions in contractual documents \citep{tuggener2020ledgar,hendrycks2021cuad}. None of these projects, however, focuses on corporate governance documents.

Substantively, the closest neighbor is CHANCERY\citep{irwin2025chanceryevaluatingcorporategovernance}, which evaluates models on corporate governance-related reasoning tasks involving real corporate charters. In CHANCERY, models are asked whether proposed executive, board, or shareholder actions are consistent with rules contained in the charter. While both projects focus on corporate governance, they differ in important respects. CHANCERY is primarily a reasoning benchmark. It tests whether a model can evaluate the permissibility of an action in light of governance rules contained in a charter. By contrast, the DECODEM Benchmarks evaluate text-to-variable translation tasks. Instead of determining the permissibility of a hypothetical action, these tasks focus on recovering structured governance variables from primary documents.

More broadly, the paper relates to the literature on information extraction in natural language processing\citep{sarawagi2008information_extraction,premasiri2025legalie}. Existing work in this area has introduced benchmarks for a variety of extraction tasks\citep{jm3,doddington-etal-2004-automatic}. Like related general-purpose NLP benchmarks\citep{wang-etal-2018-glue,wang2019superglue}, these benchmarks often focus on relatively well-defined tasks applied to shorter or more standardized texts. By contrast, the extraction tasks in DECODEM require models to process lengthy and complex legal documents and identify legally relevant arrangements that are not always apparent from the wording alone. 

The paper also relates to a rapidly growing literature examining the use of large language models for legal tasks. Recent work suggests that LLMs are capable of performing a wide range of such tasks, including legal reasoning and classification, question answering, and contract drafting (e.g., \cite{engel_mcadams_2024_gpt_ordinary_meaning,sargeant2025topic_classification_case_law,blair_stanek_etal_2023_statutory_reasoning,martin2024bettergptcomparinglarge}). More closely related to this project is work studying the ability of LLMs to extract structured information from legal sources and generate high-quality summaries of legal documents\citep{breton_etal_2025_llm_legal_terms,sym17050633}. 

Beyond the AI literature, the paper connects to the extensive law and finance literature studying the implications of corporate governance regimes (e.g., \cite{gompers2003corporate_governance,bebchuk2009what_matters}). In particular, it relates to work that evaluates the quality of existing commercial datasets \citep{frankenreiter2021cleaning_governance,LarckerReissXiao2015}, as well as to the nascent literature that recovers corporate governance provisions directly from governance documents and other filings using computational tools such as large language models \citep{guernsey2025,frankenreiter2025other_delaware_effect,frankenreiter2026sticky_charters}. Finally, the paper relates to the social science literature on the use of text as data \citep{gentzkow2019text,Grimmer_Stewart_2013}.

\section{The Benchmark Datasets}

This paper introduces two novel benchmark datasets for evaluating the automated extraction of corporate governance information from the two central organizational documents of publicly traded corporations: charters and bylaws. Both benchmarks share a common structure. They consist of a sample of documents paired with hand-coded labels capturing legally relevant governance arrangements. The charters benchmark contains 300 charters, while the bylaws benchmark contains 150 bylaws. The labels define the document-level classification tasks on which model performance is evaluated.

Unlike many benchmark datasets in natural language processing, the tasks considered here are not just of interest as measures of model capability. Instead, they correspond directly to variables that have been used (or could plausibly be used) in empirical corporate governance research. As a result, demonstrating a model's high performance on these benchmarks has immediate practical implications for corporate governance research. It validates the model as a potential part of a viable pipeline that extracts governance provisions from corporate governance documents at scale, thereby lowering the costs of constructing high-quality datasets and enabling the analysis of governance features that have previously remained outside the reach of large-scale analysis.

Previous projects that evaluated LLM-based extraction of corporate governance features did so in the context of substantive research projects\citep{frankenreiter2025other_delaware_effect,guernsey2025}. Their validation exercises therefore focused on the particular governance provisions needed for those applications. By contrast, the DECODEM benchmarks include a broad range of variables that cover different governance topics and vary in interpretive complexity. This makes it possible to evaluate systematically where automated extraction performs well and where challenges remain.

The charters and bylaws were drawn from a comprehensive corpus containing the charters and bylaws of a wide set of publicly traded corporations in the United States \citep{frankenreiter2025other_delaware_effect}. These documents were assembled through an automated pipeline using filings available on the SEC’s EDGAR database. The bylaws benchmark draws on documents filed between 1995 and 2024. For reasons related to the construction of the charters benchmark, the charters are drawn from the period between 1995 and 2019. Because both benchmarks are sampled from a broad corpus of charters and bylaws available on EDGAR, performance on these benchmarks provides a useful indication of how extraction pipelines are likely to perform when applied to the broader corpus.

The bylaws benchmark covers 25 variables, the charters benchmark 26. Most variables appear in both datasets, reflecting the fact that U.S. corporate law often allows corporate planners to implement governance arrangements in either the charter or the bylaws. Together, the benchmarks cover 31 distinct governance arrangements. Table \ref{tab:bylaws_variables} summarizes all variables and reports their incidence in both benchmarks. 

\begin{table*}[htb!]
\centering
\caption{Variables Included in the Benchmark Datasets}
\label{tab:bylaws_variables}
\small
\begin{threeparttable}
\begin{tabular}{l p{4.2cm} cc l l p{4.2cm} cc}
\toprule
\textbf{Variable} & \textbf{Description} & \multicolumn{2}{c}{\textbf{Count}}  & & \textbf{Variable} & \textbf{Description} & \multicolumn{2}{c}{\textbf{Count}} \\
 & & \textbf{BL} & \textbf{Ch} & &  & & \textbf{BL} & \textbf{Ch} \\
\midrule

\multicolumn{3}{l}{\textit{Exculpation / Indemnification}} & & &\multicolumn{3}{l}{\textit{Director Elections}} \\
IND-D & Indemnification of directors req. & 108 & 81 & & REM* & Removal only for cause. & 26 & 61 \\
IND-O & Indemnification of officers req. & 107 & 77 & & REM-S1* & Supermajority req. (for cause). & 16 & 39 \\
ADV-D & Advancement req. for directors. & 79 & 43  & & REM-S2* & Supermajority req. (w/o cause). & 7 & 16 \\
ADV-O & Advancement req. for officers. & 79 & 41 & & VAC-R* & SH vacancy authority (removal). & 25 & 11\\
INS-D & D\&O insurance req. for directors. & 6 & NA & & VAC-G* & SH vacancy authority (general). & 11 & 5 \\
INS-O & D\&O insurance req. for officers. & 6 & NA & & MAJ & Majority voting. & 35 & 11 \\
EXC-D & Director exculpation. & 3 & 153 & & PRA & Proxy access granted. & 11 & NA \\
EXC-O & Officer exculpation. & 3 & 15 & & CUMU & Cumulative voting. & NA & 8 \\
\\
\multicolumn{3}{l}{\textit{Shareholder Meetings}} & & & \multicolumn{3}{l}{\textit{Structural Features}} \\
SM-P & SHs cannot call special meetings. & 75 & 52 & & BC & Blank check preferred. & NA & 172 \\
SM-A & SHs can call special meetings. & 50 & 9 & & SMJ-C & Charter amendment smj. req. & NA & 112 \\
SM-T* & Threshold support required. & 49 & 17 & & SMJ-M & Merger supermajority req. & NA & 58 \\
WC-P & Written consent prohibited. & 32 & 83 & & MC-U* & Multiclass shares w/ uneq. votes. & NA & 27 \\
WC-A & Written consent allowed. & 50  & 16 & & TS-OO* & Anti-takeover statute opt out. & NA & 19 \\
WC-S* & Supermajority required. & 5 & 5 \\
ADJ* & Unilateral adjournment. & 49 & NA \\
\\
\multicolumn{3}{l}{\textit{Board Structure}} \\
BS-U* & Board authority to fix board size. & 60 & 36 \\
SB & Staggered board. & 40 & 76 \\
LID & Lead independent director. & 8 & NA \\

\bottomrule
\end{tabular}

\begin{tablenotes}[flushleft]
\footnotesize
\item \textit{Notes.} Variables marked with * require more complex interpretation of the document text, including distinguishing between linguistically similar but legally distinct governance arrangements or identifying conditional rules and numerical thresholds.
\end{tablenotes}

\end{threeparttable}
\end{table*}

To allow for more straightforward model evaluation, all variables take the form of binary indicators. For some governance topics, for example, the existence of a staggered board, this binary coding is straightforward. Other governance provisions present more complex coding questions. Provisions on written consent, for example, might explicitly empower shareholders to act by written consent, might contain an outright prohibition, or might make acting by written consent contingent on satisfying conditions such as a supermajority threshold. The benchmarks address this complexity by including separate variables that capture different, potentially overlapping aspects of the same governance topic. In the case of written consent, these variables are WC-P, WC-A, and WC-S.

Together, the variables define a set of extraction tasks that vary in complexity. Some variables primarily require the model to detect relatively standardized language in a document, with little legally meaningful variation across provisions. SB, which captures the existence of a staggered board, is one example. Other variables require the model to interpret the document text in more complex ways and distinguish between linguistically similar but legally distinct governance arrangements. One example is ADJ, which requires the model to determine whether the chairperson or board has unilateral authority to adjourn a shareholder meeting and whether that authority is conditioned on the absence of a quorum. While the distinction between these two groups is a matter of degree, I assign each variable to one of the two groups, which I refer to as ``Simple Variables'' and ``Complex Variables,'' respectively. Complex Variables are marked with an asterisk in Table \ref{tab:bylaws_variables}.

All variables were collected through a two-step process that combines human coding with AI-assisted review. For the bylaws benchmark, most variables were coded first by a team of research assistants, all advanced law students at U.S. law schools. Each research assistant worked on only one variable or a small set of related variables at a time. For each variable, the same research assistant coded all bylaws in the benchmark dataset. Before coding began, research assistants received detailed instructions on how to interpret the variable they were assigned. Throughout the process, they were encouraged to flag ambiguities for review. A second set of variables, including most of the variables requiring more complex interpretation, was collected by the author.\footnote{These variables are BS-U, REM-S1, REM-S2, VAC-R, and VAC-G.\label{fn:vars1}}

The information collected at this stage was more detailed than the variables ultimately included in the benchmark dataset. In particular, where this distinction made sense, coders recorded not only whether the bylaw contained the specific arrangement captured by the benchmark label, but also whether it addressed the broader governance topic at all. For example, one question asked coders to determine whether the bylaws contained any provision governing director removal. A second question then asked whether director removal was limited to removal for cause only. In addition, for all variables, coders were asked to quote the text on which they based their coding decision.

For the charters benchmark, most variables were drawn from a previous project that collected detailed information on the contents of charters with the help of a larger team of research assistants\cite{frankenreiter2021cleaning_governance}. A small set of additional variables was again collected by the author.\footnote{This set includes the same variables mentioned in note \ref{fn:vars1} plus TS-OO.}

In a second step, the initial codings were subjected to an AI-assisted audit. The design of this audit differed depending on whether the available information indicated the presence or absence of language related to the relevant provision. When the data indicated the presence of a relevant provision type, GPT-5.2 was prompted to review the corresponding coding decisions for individual variable categories---for example, the two questions related to director removal discussed above---against the text quoted by the human coder. Cases that GPT-5.2 flagged as unsupported were reviewed either by the author or by a senior research assistant at Washington University in St. Louis, working under close supervision.

When the information collected in the first step indicated the absence of a relevant provision type, GPT-5-mini was prompted to review the full document text for paragraphs containing language related to the relevant governance topic. 
In instances where this model flagged potentially relevant language, GPT-5.2 was then prompted to assess whether the identified passages in fact contained the provision at issue. 
Cases in which GPT-5.2 identified the provision as present were reviewed by the author, who corrected the coding where necessary. 

Two features of this process deserve to be mentioned. First, all final labeling decisions were made by humans. Automated methods were used only to flag coding decisions for review. In no case did an LLM produce the final labels used to evaluate model performance. Second, the prompts used for validation were developed separately from the extraction routines evaluated in the following section, and the audit relied on earlier-generation models than those evaluated below. Together, these design choices reduce the risk that the benchmark labels reproduce the same blind spots and biases as the models or prompts evaluated later.

Companion work extends this validation process one step further \citep{frankenreiter2026measuring}. Every observation for which there is less than perfect agreement among the initial human coding, the benchmark label, and six extraction routines described below is recoded blindly, without knowledge of the existing labels. This recoding upholds the benchmark labels in a large majority of reviewed cases, ultimately revising less than 1\% of observations in the charters benchmark and around 1.5\% of observations in the bylaws benchmark. This process produces an enhanced set of reference labels (the ``Adjudicated Benchmarks''), which incorporate the revised labels for observations where the original benchmark label was overturned. Because the Adjudicated Benchmarks were constructed only after observing disagreements between the DECODEM Benchmarks and some of the extraction routines evaluated here, the present paper evaluates model and pipeline performance against the original DECODEM Benchmarks rather than the Adjudicated Benchmarks. This avoids biasing the results in favor of extraction routines whose decisions were taken into account when constructing the latter benchmarks. That said, because some adjudicated revisions resolved model-benchmark disagreements in favor of the extraction routines, the performance estimates reported below are likely somewhat conservative.

The DECODEM benchmarks, the Adjudicated Benchmarks, the underlying charter and bylaw documents, the extraction prompts and pipelines, and the evaluation code will be made publicly available upon publication under a permissive license. Materials are available from the author on reasonable request prior to publication.
\section{Extraction Approaches and Models}

This paper evaluates the performance of several automated extraction approaches. The goal is not only to identify which method performs best overall, but also to shed light on which pipeline design choices are associated with stronger extraction performance.

I evaluate several extraction pipelines that vary along three dimensions. First, the pipelines vary in whether the model is provided with the full document or only selected excerpts. These excerpts are generated using a BERT- and LASSO-based pipeline that predicts, for each paragraph, whether it contains language relevant to a given governance provision (for details, see \cite{frankenreiter2025other_delaware_effect}). Because the model used for extraction should be able to tolerate irrelevant passages, the excerpting step is designed to prioritize recall over precision. The main purpose of this approach is to reduce extraction costs by limiting the number of tokens submitted to the extraction model (see \cite{li2023cascading}). To the extent that model performance is sensitive to long context windows, this ``cascading pipeline'' may also improve extraction accuracy.

Second, the pipelines vary in prompt complexity and, in particular, in the amount of task-specific guidance the prompts provide. The first set of approaches relies on relatively simple prompts that ask the model whether a given governance arrangement is present in the text. These prompts do not provide detailed definitions of the relevant governance provisions or examples of how they may be expressed. They therefore require relatively little task-specific prompt engineering and avoid the need to encode detailed domain knowledge directly in the prompt. As a result, these prompts rely primarily on the model’s existing understanding of corporate governance concepts.

The second set of approaches relies on more detailed prompts that provide the model with background information and explicit rules for resolving harder cases. Unlike the simple prompts described above, these prompts do not rely primarily on the model’s latent understanding of corporate governance concepts. Instead, they supply detailed instructions intended to guide the model's decisions and enhance consistency in ambiguous cases. Designing these prompts therefore requires substantially more domain-specific knowledge than designing the simple prompts. Appendix \ref{app:pipelines} contains examples from both families of prompts.

\begin{table*}[htb!]
\centering
\caption{Extraction Approaches and Models}
\label{tab:extraction_design}
\small
\begin{threeparttable}

\begin{tabular}{l l l l}
\toprule
\multicolumn{4}{l}{\textit{Panel A: Extraction Approaches}} \\
\midrule
\textbf{Approach} & \textbf{Prompts} & \textbf{Document Input} & \textbf{Extraction Scope} \\
\midrule
Global
    & Simple
    & Full document
    & All variables in a single run \\

Simple Full
    & Simple
    & Full document
    & Variables related to a specific topic \\

Simple Cascade
    & Simple
    & Selected excerpts
    & Variables related to a specific topic \\

Detailed Full
    & Detailed
    & Full document
    & Variables related to a specific topic \\

Detailed Cascade
    & Detailed
    & Selected excerpts
    & Variables related to a specific topic \\

\midrule
\multicolumn{4}{l}{\textit{Panel B: Models}} \\
\midrule
\textbf{Model} & \textbf{Provider} & \textbf{Type} & \textbf{Snapshot} \\
\midrule

GPT-5.4
    & OpenAI
    & Frontier
    & \texttt{gpt-5.4-2026-03-05} \\

GPT-5.4-mini
    & OpenAI
    & Efficiency
    & \texttt{gpt-5.4-mini-2026-03-17} \\

Claude Opus 4.7
    & Anthropic
    & Frontier
    & \texttt{claude-opus-4-7} \\

Claude Sonnet 4.6
    & Anthropic
    & Efficiency
    & \texttt{claude-sonnet-4-6} \\

Gemini Pro 2.5
    & Google
    & Frontier
    & \texttt{gemini-2.5-pro} \\

Gemini Flash 2.5
    & Google
    & Efficiency
    & \texttt{gemini-2.5-flash} \\

\bottomrule
\end{tabular}

\end{threeparttable}
\end{table*}

Finally, the pipelines vary in whether variables are extracted in one or multiple runs. Most approaches extract variables over multiple runs, each covering one set of related variables. One pipeline instead asks the model to extract all variables in a single run. This approach is feasible only when the model receives the full document text, and I only implement it for the simple prompts to avoid constructing an excessively long and complex prompt. Overall, this setup yields five extraction approaches, detailed in Panel A of Table \ref{tab:extraction_design}.\footnote{All prompts required models to return outputs conforming to a fixed JSON schema that encoded the benchmark variables in a structured format that could be parsed automatically.}

During development, outputs for individual benchmark documents were inspected to check that the pipelines performed the intended extraction tasks. All approaches were finalized before full benchmark results were available.

In addition to the extraction approach, I vary the model used for the task. I focus on six recent frontier and efficiency-oriented models from the three major AI labs: GPT-5.4, GPT-5.4-mini, Claude Opus 4.7, Claude Sonnet 4.6, Gemini Pro 2.5, and Gemini Flash 2.5.\footnote{These were the most recent models available as stable releases when model selection was finalized in April 2026; in particular, Gemini 3 models were available only as previews at the time.} All models were accessed through their respective provider APIs between March 27, 2026 and June 19, 2026 using the snapshots detailed in Panel B of Table \ref{tab:extraction_design}.\footnote{Repeated runs for the principal models and approaches produced nearly identical aggregate results. The following sections therefore only report results from the first run. Appendix \ref{app:repeated-runs} reports these robustness checks.} For models that allow users to set temperature, I used a temperature of 0. I left all other parameters at provider defaults.\footnote{Outputs that could not be parsed or omitted a required field were discarded and the prompt resubmitted under the same settings until a valid JSON object was returned. Outputs were discarded only for formatting failures and not in response to the substantive content of a prediction.}

\section{Extraction Performance--Bylaws}

This section reports the performance of the models and extraction approaches detailed in Section 5 above on the Bylaws Benchmark. Here and in the remainder of the paper, performance is evaluated by comparing model-generated labels to the DECODEM labels described in Section 4.\footnote{Appendix \ref{app:adjudicated} reports results for the Adjudicated Benchmark.}

\begin{figure*}[htb!]
\centering
\includegraphics[width=\textwidth]{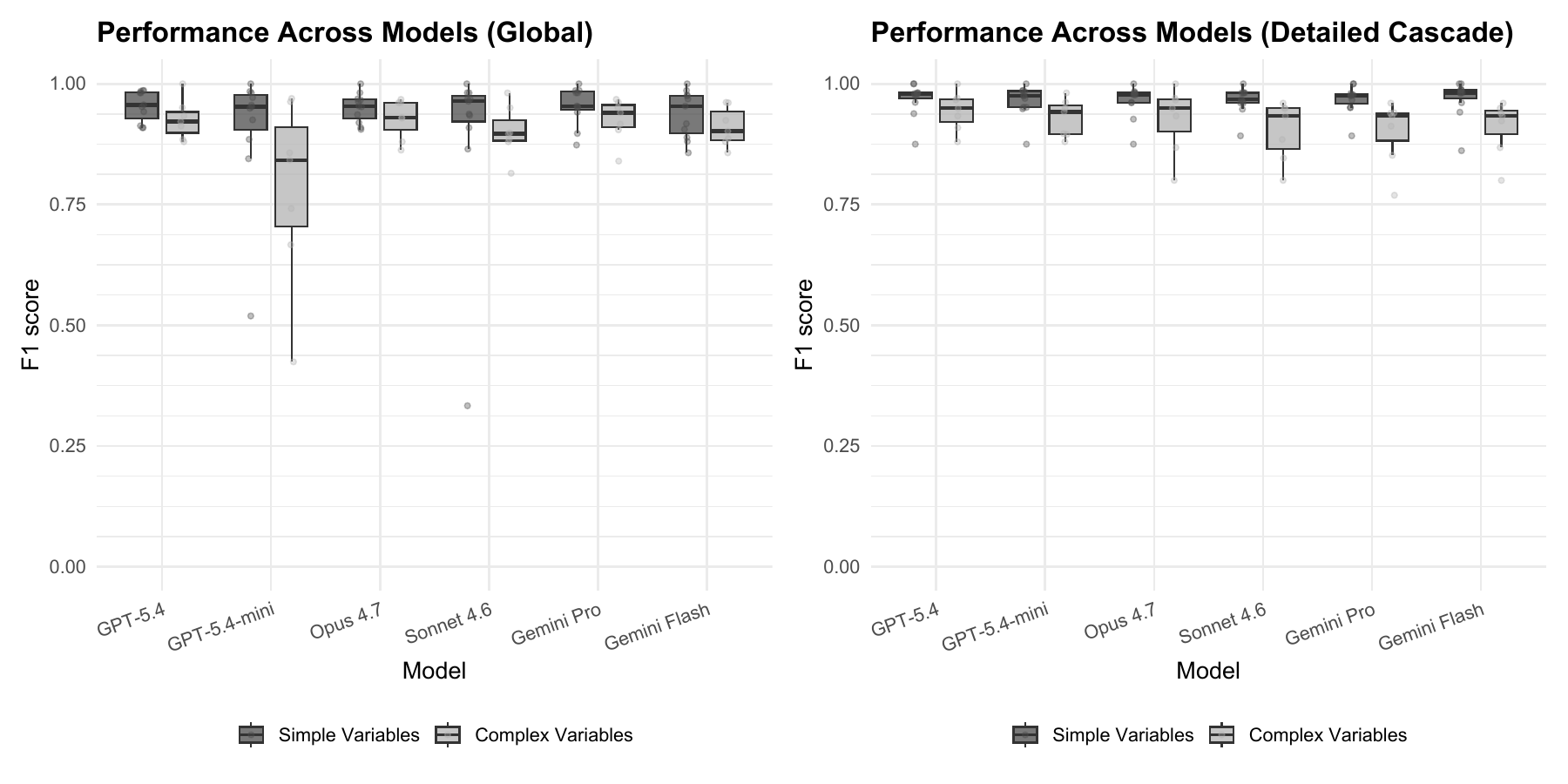}
\caption{Variable-level F1 scores for extraction of bylaw features across models.}
\label{fig:boxplot_flagship}
\end{figure*}

\paragraph{Model comparisons.} Figure \ref{fig:boxplot_flagship} compares performance across all models for two extraction approaches: Global (left column) and Detailed Cascade (right column). These approaches sit on two opposite sides of the spectrum of approaches included in this study. Global uses simple prompts and involves no preprocessing of the bylaws. Detailed Cascade uses the detailed prompts and relies on excerpting parts of a bylaw before the text is fed into the model. Within each approach, the darker box plots display the Simple Variables, while the lighter box plots represent the Complex Variables. Because F1 scores can be unstable for variables with very few positive instances, Figure \ref{fig:boxplot_flagship} (as well as Figures \ref{fig:boxplot_highlevel} and \ref{fig:boxplot_flagship_charters} below) includes only variables with Count $\geq 10$.

For the Simple Variables, performance is uniformly strong, with median F1 scores close to one under both the Global and Detailed Cascade approaches. Complex Variables exhibit substantially greater variation.

Differences also emerge across models. Within each model family, the frontier model appears to outperform the more efficiency-oriented one. This is particularly visible for the GPT family under the Global approach, where GPT-5.4-mini appears to perform noticeably worse than GPT-5.4. By contrast, there appears to be relatively little variation among the three frontier models under both approaches.

I use paired bootstrap confidence intervals to assess how stable the observed differences are. I draw 10,000 document-level bootstrap samples and focus on differences in macro-averaged F1 between models.\footnote{Appendix \ref{app:bootstrapping} describes the bootstrap procedure and reports the pairwise comparisons in detail.}

I start with differences between frontier and efficiency-oriented models from the same model family. Under the Global approach, there is a systematic gap. In each model family, the frontier model outperforms its efficiency-oriented counterpart, with confidence intervals excluding zero. Under Detailed Cascade, this gap largely disappears. The confidence intervals for all comparisons include zero, and for Gemini, Flash appears to beat Pro. This pattern suggests an interaction between model capability and extraction architecture.

This interaction is also visible when comparing performance under Global and Detailed Cascade for each model. For the frontier models, Global and Detailed Cascade yield broadly similar overall performance. Confidence intervals include zero for all three frontier models. By contrast, Detailed Cascade substantially improves performance for the three efficiency-oriented models. The bootstrap results place the gain at roughly 7 to 12 F1 points for GPT-5.4-mini, 2 to 6 for Sonnet 4.6, and 0.2 to 4 for Gemini Flash.

Finally, the interaction is confirmed by the corresponding difference-in-differences estimates. These estimates compare each efficiency-oriented model's gain from Detailed Cascade to the gain for its frontier counterpart. Confidence intervals exclude zero in all three families. This implies that the more elaborate pipeline mainly benefits the efficiency-oriented models. Frontier models, by contrast, appear able to achieve similar performance without the additional guidance and preprocessing provided by the more elaborate pipeline.

Among the three frontier models, pairwise differences are generally small. Under the Global approach, all three confidence intervals include zero and are no wider than roughly 2.5 F1 points. Under Detailed Cascade, GPT-5.4 outperforms Gemini Pro 2.5 by a small margin, with a confidence interval stretching from 0.6 to 3 F1 points. The confidence intervals for the other comparisons include zero here as well. Overall, the results suggest that the three frontier models are largely interchangeable for the task considered here. There is only one comparison that excludes zero, its lower bound is close to zero, and the largest gap is only about two F1 points.

Overall, these results are notable. Even using simple prompts, current frontier models appear capable of extracting legally relevant information from corporate bylaws with a high degree of accuracy. At the same time, the figure shows that the lower tail in performance is driven almost entirely by variables requiring more complex interpretation. These results suggest that the main challenge is not average performance---which is near ceiling across all models---but variation across variables.\footnote{Appendix \ref{app:errors} analyzes overlap between the errors made by different models. The main finding is that these errors overlap only weakly, especially under the Global approach. The relationship between classifier error diversity and the potential gains from aggregation has long been studied in the ensemble-learning literature\citep{dietterich2000,kuncheva2003mlj-measures}. Companion work builds on this finding to develop ensemble approaches that combine codings obtained from different extraction routines \citep{frankenreiter2026measuring}.}

\paragraph{Extraction approaches.} Having compared performance across models, I now turn in more detail to variation across extraction architectures. Figure \ref{fig:boxplot_highlevel} compares all five extraction architectures using GPT-5.4 and Opus 4.7, two of the strongest performers for the task evaluated here. 

\begin{figure*}[htb!]
\centering
\includegraphics[width=\textwidth]{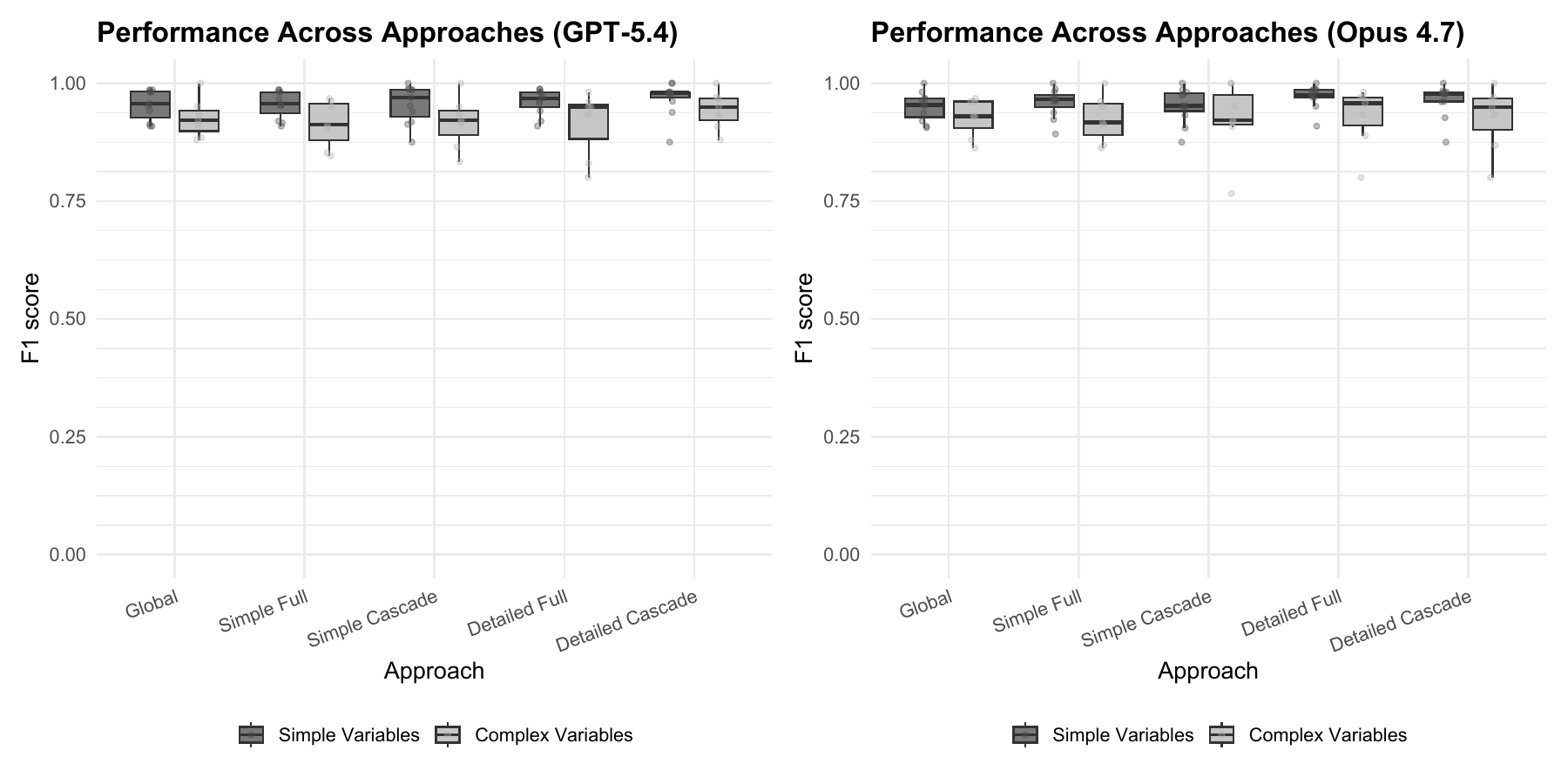}
\caption{Variable-level F1 scores for extraction of bylaw features across extraction approaches.}
\label{fig:boxplot_highlevel}
\end{figure*}

Consistent with the results above, differences across approaches remain modest overall. Simple Full, Simple Cascade, and Detailed Full exhibit performance levels roughly in line with Global and Detailed Cascade. Neither cascading extraction nor more detailed prompting appears to consistently improve performance.

To compare the approaches more directly, I again use paired bootstrap confidence intervals, this time holding the model fixed. For GPT-5.4, performance differs little across the five approaches, which all fall within roughly two F1 points of one another. Detailed Cascade nevertheless outperforms Simple Full, Simple Cascade, and Detailed Full by small margins, with confidence intervals excluding zero. For the four multi-run approaches, I separately examine the effects of detailed prompting and excerpting. The confidence intervals for detailed prompting, excerpting, and the interaction between the two all include zero. The same is true when comparing the Global and Simple Full approaches, which are equivalent except for the fact that Global extracts all variables in a single pass. If anything, this comparison slightly favors Global. Repeating the same analyses for Claude Opus 4.7 produces the same overall pattern.

Together, these results lend additional support to the conclusion that current frontier models perform similarly across prompt design, document handling, and task decomposition choices. In particular, the fact that cascading approaches do not consistently outperform approaches that provide models with the full document suggests that current frontier models can often process long legal documents without document size negatively affecting performance.\footnote{Appendix \ref{app:variable-level} provides a deeper analysis of variable-level results, comparing the performance of different variables both across models and approaches.} This result is noteworthy given the substantial length of bylaw documents, which often exceed 10,000 words and sometimes surpass 20,000.\footnote{Appendix \ref{app:doclength} investigates the relationship between document length and extraction performance in more detail.}

\section{Extraction Performance--Charters}

This section extends the analysis to the charters benchmark, testing whether the findings reported for bylaws generalize to a second class of document. Since I rely on the same extraction pipelines and models, results are directly comparable across the two document types.

Charters differ from bylaws in several ways. They are often shorter, which might facilitate extraction. At the same time, their structure can vary substantially, in particular because charters may contain detailed provisions governing preferred stock, posing a potential challenge for extraction. 

\begin{figure*}[htb!]
\centering
\includegraphics[width=\textwidth]{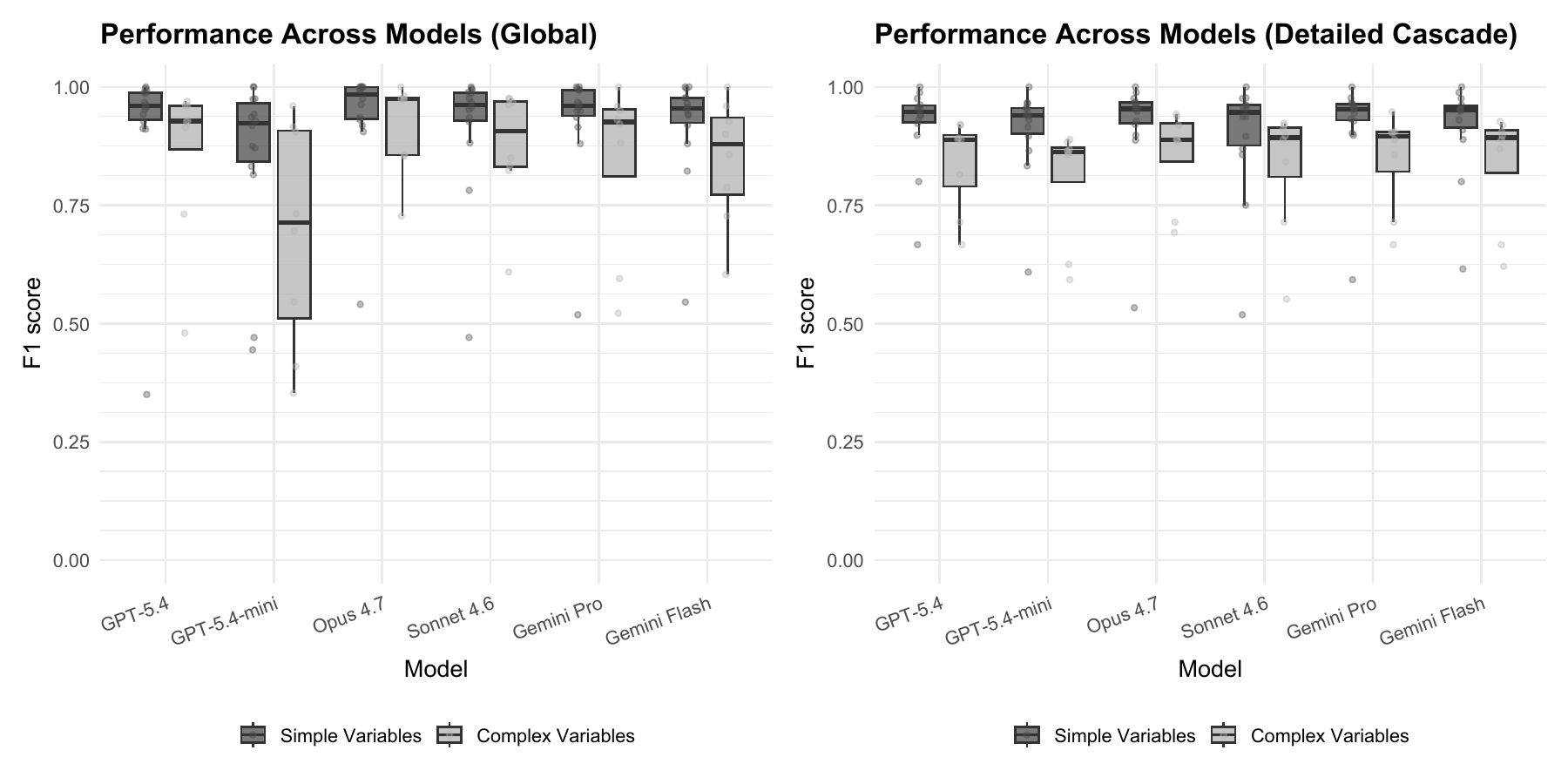}
\caption{Variable-level F1 scores for extraction of charter features across models.}
\label{fig:boxplot_flagship_charters}
\end{figure*}

Figure \ref{fig:boxplot_flagship_charters} replicates Figure \ref{fig:boxplot_flagship}, reporting variable-level F1 scores across the six models for the Global and Detailed Cascade approaches. As before, this figure excludes variables with a count of $< 10$. The overall picture broadly mirrors the bylaws results. For the Simple Variables, performance is uniformly strong, with median F1 scores clustered near one under both approaches. The Complex Variables exhibit substantially greater dispersion. Within each model family, the larger models tend to outperform their efficiency-oriented counterparts on the Complex Variables, a gap most visible for GPT-5.4-mini under the Global approach. 

Bootstrap comparisons, however, reveal two departures from the bylaws benchmark. First, the frontier models are less interchangeable here. Under the Global approach, Opus 4.7 outperforms the other two with confidence intervals excluding zero. Yet the same result cannot be observed under Detailed Cascade. Second, the interaction between model type and approach observed for the bylaws benchmark only replicates for the GPT model family. More broadly, only GPT-5.4-mini gains systematically under Detailed Cascade. Some models, notably Opus 4.7, perform worse under Detailed Cascade than under Global.

Furthermore, comparing Figures \ref{fig:boxplot_flagship_charters} and \ref{fig:boxplot_flagship} suggests that average performance on the charters benchmark trails that on the bylaws benchmark.

\begin{figure*}[htb!]
\centering
\includegraphics[width=\textwidth]{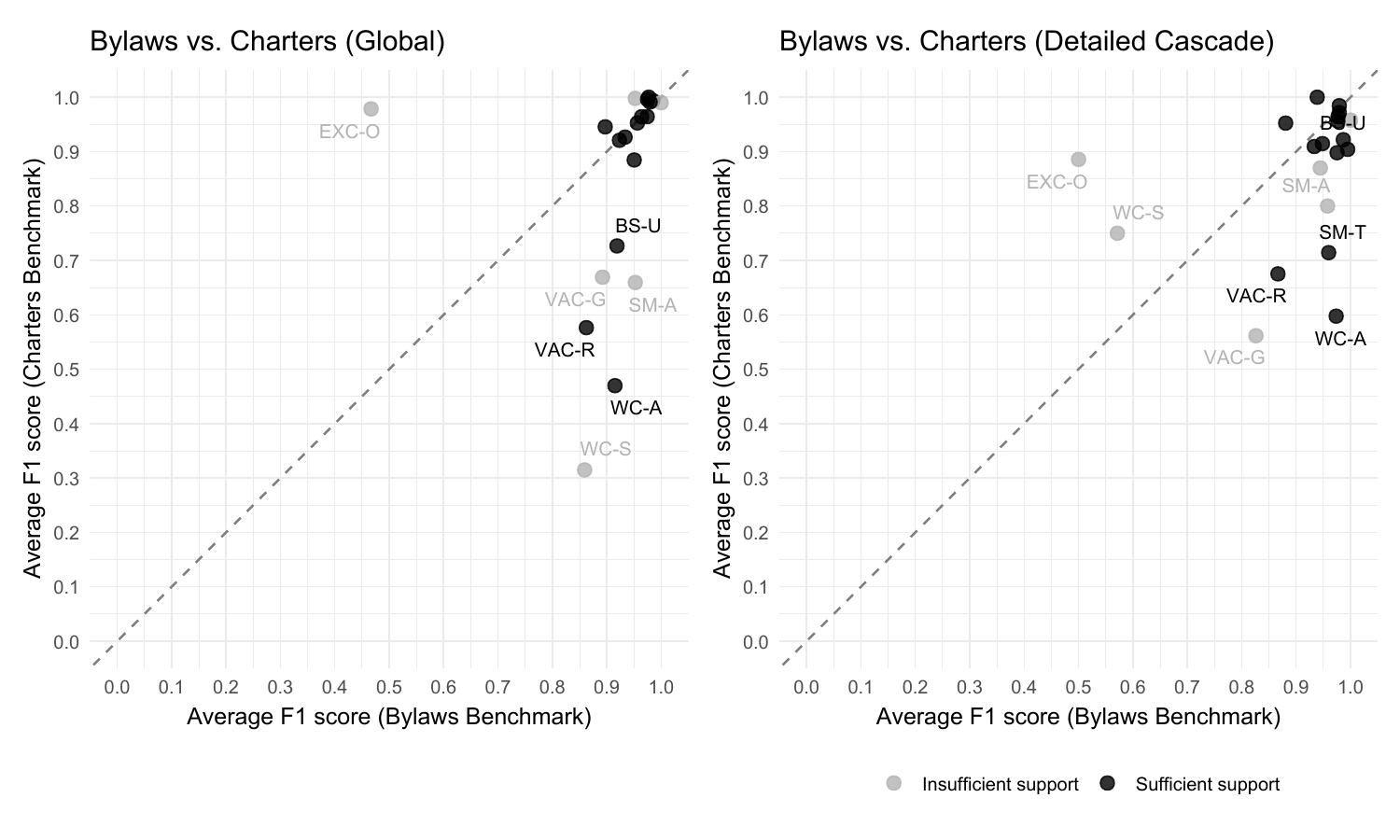}
\caption{Comparison of performance for variables featured in both benchmarks. Average F1 scores are computed across GPT-5.4, Claude Opus 4.7, and Gemini Pro 2.5.}
\label{fig:chartersvsbylaws}
\end{figure*}

Figure \ref{fig:chartersvsbylaws} sheds more light on this gap by comparing the performance on variables that appear in both benchmarks. Because variables are present in both benchmarks and are extracted using identical prompts, this figure represents document-type effects as opposed to effects related to variable selection and prompt design. Systematic underperformance on charters would suggest that some structural feature of charter documents makes legally relevant information harder to recover. The figure plots average F1 for the three frontier models on the charters benchmark (y-axis) against average F1 on the bylaws benchmark (x-axis) under the Global and Detailed Cascade approaches. The dashed 45-degree line marks equal performance on both benchmarks. Points below the line indicate that extraction performance from charters trails extraction performance from bylaws for a given variable. Colors distinguish variables with $\geq$ 10 observations in both benchmarks (those variables included in Figures \ref{fig:boxplot_flagship} and \ref{fig:boxplot_flagship_charters}) from variables with fewer observations. 

Most variables cluster in the upper-right corner, close to the dashed line. This suggests that, for many variables, extraction performance is stable across document types. Provisions that are easy to recover from bylaws are, for the most part, equally easy to recover from charters. 

Yet a few variables depart from this pattern. The first group of variables with a different pattern belongs to the variables with low N in at least one of the benchmarks. EXC-O, for example, lies above the line, reflecting better performance on charters than on bylaws. Yet the bylaws benchmark contains only three positive instances, so its F1 score is highly sensitive to individual errors, whereas the larger number of positive instances in the charters benchmark yields a more stable estimate. That said, one possible explanation for the discrepancy is that models more readily ``expect'' to encounter officer-exculpation provisions in charters than in bylaws, but this possibility is impossible to verify given the low count of true ``Ys'' for bylaws.

Another group of variables consists of variables with sufficient support in both benchmarks. These variables include VAC-R and various variables from the WC and SM categories. All these variables exhibit stronger performance in the bylaws benchmark. This difference in performance might be due to the distorting effect of provisions governing preferred stock, which sometimes contain provisions applicable exclusively to the holders of specific series of preferred stock. Those provisions were excluded from the definitions used to assemble the benchmarks but likely confuse the extraction algorithms in some instances.

To corroborate the document-type gap visible in Figure \ref{fig:chartersvsbylaws}, I again compute bootstrap confidence intervals. Here, the intervals represent the difference in macro-averaged F1 between bylaws and charters on variables with at least ten positive instances in both benchmarks. Across the Global and Detailed Cascade approaches and the three frontier models, bylaws outperform charters by roughly 3 to 8 F1 points, with all confidence intervals excluding zero. This finding supports the conclusion that there are systematic differences associated with extracting the same governance variables from the two document types, at least for the prompts, models, and extraction approaches used here.

In sum, the charters results support the central findings of the bylaws analysis. Automated extraction performs well for the large majority of governance provisions, including several structural and anti-takeover provisions that do not appear in the bylaws benchmark. Performance again varies systematically across variables, with more complex provisions accounting for most of the remaining errors. As before, more detailed prompting and cascading pipelines do not deliver consistent gains for frontier models. The close correspondence between the two benchmarks shown in Figure \ref{fig:chartersvsbylaws} provides additional evidence that these findings are broadly consistent across document types. There are, however, a few notable exceptions for which performance on charters lags performance on bylaws. These differences are concentrated in shareholder-meeting, written-consent, and vacancy provisions and plausibly reflect the presence of preferred-stock provisions in charters.
\section{Limitations}

Several limitations apply. Most importantly, the benchmark labels are not ground truth. The AI-assisted audit described in Section 4 was designed to surface and correct errors in the initial human codings, but it does not guarantee that the labels are error-free. The blind re-coding of contested observations in \citet{frankenreiter2026measuring} confirms the existence of some residual noise, but also suggests that it is limited: it revises less than 1\% of charter observations and around 1.5\% of bylaw observations. Moreover, the robustness checks reported in Appendix \ref{app:adjudicated} suggest that the main results persist if the Adjudicated Benchmarks are used in place of the DECODEM Benchmarks; if anything, the accuracy estimates based on the DECODEM Benchmarks are likely conservative.

Second, the benchmarks necessarily reflect choices about what variables are included and how each variable is defined. Those choices can affect measured performance. For example, provisions that apply only to holders of particular series of preferred stock are excluded from the variable definitions. As discussed in Section 7, this likely contributes to weaker performance on some shareholder-meeting and written-consent variables in the charters benchmark. More generally, a variable may appear easier or harder depending on how its boundaries are drawn, and different but reasonable coding rules could produce somewhat different results. Similarly, a different set of variables might have yielded different results, although the inclusion of a broad range of variables with varying levels of complexity reduces this concern.

Third, the benchmarks cover a particular set of documents and variables. The documents are drawn from SEC EDGAR filings for U.S. publicly traded corporations \citep{frankenreiter2025other_delaware_effect}. The bylaws span filings between 1995 and 2024, while the charters cover 1995 through 2019. The results may not hold for other firms or time periods. 

A fourth concern is possible data contamination. The charters and bylaws used here are publicly available on EDGAR and may have featured in the models' pretraining data. It is therefore possible that the measured accuracy partly reflects the models' familiarity with specific documents rather than a general capacity to extract governance provisions. That said, the tasks evaluated here require more than familiarity with the documents. They require the model to extract variables according to specific coding rules, and, with the exception of some variables in the charters benchmark, no publicly available dataset links these documents to the variables evaluated here. Still, this concern cannot be ruled out completely. The results should thus be interpreted as evidence about performance on previously published filings similar to those included in the benchmarks. Performance on novel filings could differ.

Finally, assembling the DECODEM Benchmarks relied on an AI-assisted audit using earlier OpenAI models (GPT-5.2 and GPT-5-mini), while more recent models from the same family were included in the analysis. This raises the possibility that OpenAI models may have benefited from what one might call a same-family advantage. The fact that human coders set all final labels and that the validation prompts were developed separately from the extraction routines reduces this concern. However, as with data contamination, it cannot be ruled out completely.
\section{Conclusion}

This paper introduces the DECODEM Benchmarks and uses them to evaluate whether large language models can extract corporate governance variables from charters and bylaws and to compare performance across models and extraction approaches. Overall, the results are encouraging. For many governance provisions, current models achieve high accuracy levels. At the same time, performance varies substantially across documents and variables. Most remaining errors are concentrated in a relatively small number of provisions that are particularly difficult to extract from charters. By contrast, most standardized provisions, such as indemnification and insurance provisions, can generally be extracted with very high accuracy across documents.

The results also show that frontier models perform comparably on the tasks evaluated here and that more elaborate extraction pipelines do not consistently improve performance. For these models, approaches incorporating detailed prompts and cascading pipelines perform broadly similarly to simpler approaches. More elaborate extraction strategies matter more for efficiency-oriented models, where they can substantially narrow the performance gap relative to frontier models.

These findings have important implications for empirical corporate governance research. Automated extraction can make it possible to study governance arrangements that have previously been difficult to analyze at scale. In particular, if governance variables can be extracted reliably from primary documents, researchers no longer face the tradeoff between expensive hand coding and relying on the limited set of variables available in existing datasets. At the same time, without validation, it remains difficult to know which variables can be extracted reliably.

\label{lastpage}

\bibliographystyle{ACM-Reference-Format}
\bibliography{paper}

\clearpage
\onecolumn

\begingroup
\normalsize
\onehalfspacing

\appendix

\counterwithin{figure}{section}
\counterwithin{table}{section}

\section{Bootstrapped Comparisons}
\label{app:bootstrapping}

This Appendix reports the methodology as well as detailed results for the bootstrapped
comparisons discussed in Sections 6 and 7. 

As discussed in the main text, the comparisons are based on macro-averaged F1 scores. I obtain these scores by computing variable-level F1 scores and averaging them across all variables with at least ten positive instances in the original benchmark (18 for the bylaws benchmark, 22 for the charter benchmark). Each comparison is based on estimates from the original data together with confidence intervals derived from 10,000 bootstrap samples. I sample at the document level in order to account for within-document dependencies, which are particularly relevant given that some variables cover related corporate governance concepts. 

Table \ref{tab:bootstrap_bylaws} reports the comparisons underlying the results reported in the \textit{Model comparisons} part of Section 6. Panel A compares the frontier and efficiency-oriented models within each model family. Under the Global approach, frontier models outperform their efficiency-oriented counterparts in all three families. All confidence intervals exclude zero. Under Detailed Cascade, these differences become much smaller, and none of the three confidence intervals excludes zero. 

\begin{table*}[h]
\centering
\caption{Bootstrap Comparisons of Model Performance--Bylaws}
\label{tab:bootstrap_bylaws}
\small

\begin{threeparttable}

\begin{tabular}{l c c c c}
\toprule
\multicolumn{5}{l}{\textbf{Panel A: Frontier vs. Efficiency Models}} \\
\midrule
\textbf{Model Comparison} &
\textbf{Model 1} &
\textbf{Model 2} & 
\textbf{Difference} &
\textbf{95\% CI} \\
\midrule

\multicolumn{5}{l}{\textit{Global}} \\
GPT-5.4 -- GPT-5.4-mini            & 0.944 & 0.857 & 0.086 & [0.063, 0.113] \\
Opus 4.7 -- Sonnet 4.6            & 0.942 & 0.897 & 0.044 & [0.029, 0.060] \\
Gemini Pro 2.5 -- Gemini Flash 2.5 & 0.944 & 0.927 & 0.017 & [0.004, 0.031] \\

\addlinespace
\multicolumn{5}{l}{\textit{Detailed Cascade}} \\
GPT-5.4 -- GPT-5.4-mini            & 0.959 & 0.950 & 0.008 & [-0.004, 0.021] \\
Opus 4.7 -- Sonnet 4.6            & 0.950 & 0.941 & 0.008 & [-0.008, 0.025] \\
Gemini Pro 2.5 -- Gemini Flash 2.5 & 0.942 & 0.947 & -0.005 & [-0.014, 0.003] \\

\midrule

\multicolumn{5}{l}{\textbf{Panel B: Detailed Cascade vs. Global}} \\
\midrule
\textbf{Model} &
\textbf{Cascade} &
\textbf{Global} &
\textbf{Difference} &
\textbf{95\% CI} \\
\midrule

\multicolumn{5}{l}{\textit{Frontier}} \\
GPT-5.4        & 0.959 & 0.944 & 0.015  & [-0.0001, 0.031] \\
Opus 4.7       & 0.950 & 0.942 & 0.008  & [-0.007, 0.022] \\
Gemini Pro 2.5 & 0.942 & 0.944 & -0.002 & [-0.020, 0.014] \\

\addlinespace
\multicolumn{5}{l}{\textit{Efficiency}} \\
GPT-5.4-mini     & 0.950 & 0.857 & 0.093 & [0.071, 0.119] \\
Sonnet 4.6       & 0.941 & 0.897 & 0.044 & [0.024, 0.063] \\
Gemini Flash 2.5 & 0.947 & 0.927 & 0.020 & [0.002, 0.037] \\

\addlinespace
\multicolumn{5}{l}{\textit{Difference-in-Differences}} \\
GPT-5.4 -- GPT-5.4-mini            & 0.008 & 0.086 & -0.078 & [-0.106, -0.052] \\
Opus 4.7 -- Sonnet 4.6            & 0.008 & 0.044 & -0.036 & [-0.059, -0.014] \\
Gemini Pro 2.5 -- Gemini Flash 2.5 & -0.005 & 0.017 & -0.022 & [-0.039, -0.006] \\

\midrule

\multicolumn{5}{l}{\textbf{Panel C: Frontier Model Comparisons}} \\
\midrule
\textbf{Model Comparison} &
\textbf{Model 1} &
\textbf{Model 2} &
\textbf{Difference} &
\textbf{95\% CI} \\
\midrule

\multicolumn{5}{l}{\textit{Global}} \\
GPT-5.4 -- Opus 4.7        & 0.944 & 0.942 & 0.002  & [-0.009, 0.013] \\
GPT-5.4 -- Gemini Pro 2.5  & 0.944 & 0.944 & 0.000  & [-0.012, 0.012] \\
Opus 4.7 -- Gemini Pro 2.5 & 0.942 & 0.944 & -0.002 & [-0.013, 0.008] \\

\addlinespace
\multicolumn{5}{l}{\textit{Detailed Cascade}} \\
GPT-5.4 -- Opus 4.7        & 0.959 & 0.950 & 0.009 & [-0.006, 0.025] \\
GPT-5.4 -- Gemini Pro 2.5  & 0.959 & 0.942 & 0.017 & [0.006, 0.030] \\
Opus 4.7 -- Gemini Pro 2.5 & 0.950 & 0.942 & 0.008 & [-0.007, 0.023] \\

\bottomrule
\end{tabular}

\begin{tablenotes}[flushleft]
\footnotesize
\item \textit{Notes.} Entries report macro-averaged F1 scores across the 18 variables with at least ten positive instances in the bylaws benchmark. In Panel A, Difference is the frontier-model F1 score minus the corresponding efficiency-model F1 score. In Panel B, Difference is the Detailed Cascade F1 score minus the Global F1 score. The Difference-in-Differences rows report, for each model family, the performance gap between frontier and efficiency models under both Detailed Cascade and Global; Diff-in-Diff is the Detailed Cascade gap minus the Global gap. In Panel C, Difference is the first model's F1 score minus the second model's F1 score. Confidence intervals are percentile-based 95\% confidence intervals from document-level bootstraps with $R=10,000$ iterations.
\end{tablenotes}

\end{threeparttable}
\end{table*}

Panel B reports within-model comparisons between Detailed Cascade and Global. For the efficiency models, particularly for GPT-5.4-mini and Sonnet 4.6, Detailed Cascade produces large gains as compared to Global. By contrast, the gains for the frontier models are much smaller; all confidence intervals include zero. The difference-in-differences estimates
in the same panel confirm the existence of a systematic difference in how different extraction approaches affect the performance of more capable vs. more efficient models: the frontier-efficiency gap narrows substantially under Detailed Cascade in all three model families, and the confidence intervals for this effect exclude zero for all three model families.

Panel C compares the three frontier models directly. Differences between all models are below two F1 points under both approaches, and substantially below that under Global. Only the GPT-5.4--Gemini Pro 2.5 comparison under Detailed Cascade yields a confidence interval excluding zero.

Table \ref{tab:bootstrap_bylaws2} reports the comparisons underlying the results reported in the \textit{Extraction approaches} part of Section 6. Panel A reports pairwise comparisons between the five extraction approaches, separately for GPT-5.4 and Opus 4.7. Consistent with the main results, most differences are small, and most confidence intervals include zero. The only exceptions are the comparisons of Detailed Cascade with Simple Full, Simple Cascade, and Detailed Full under GPT-5.4. Detailed Cascade performs better in all three comparisons, and the corresponding confidence intervals exclude zero. For Opus 4.7, by contrast, none of the pairwise differences is statistically significant.

Panel B aggregates the four approaches that use multiple runs (all approaches except Global) along the two design dimensions on which they vary: document handling and prompt detail. For both models, neither excerpting nor more detailed prompts produces a statistically significant improvement, and the interaction between the two design choices is likewise not statistically significant. Together with the finding that Global and Simple Full show roughly similar performance, these results lend additional support to the conclusion that no single extraction-design choice consistently improves performance for the frontier models.

\begin{table*}[h]
\centering
\caption{Bootstrap Comparison of Approach Performance--Bylaws}
\label{tab:bootstrap_bylaws2}
\small

\begin{threeparttable}

\begin{tabular}{l c c c c}
\toprule
\multicolumn{5}{l}{\textbf{Panel A: Extraction Approaches}} \\
\midrule
\textbf{Extraction Approaches} &
\textbf{Approach 1} &
\textbf{Approach 2} &
\textbf{Difference} &
\textbf{95\% CI} \\
\midrule

\multicolumn{5}{l}{\textit{GPT-5.4}} \\
Global -- Simple Full              & 0.944 & 0.939 & 0.004  & [-0.006, 0.015] \\
Global -- Simple Cascade           & 0.944 & 0.941 & 0.003  & [-0.010, 0.014] \\
Global -- Detailed Full            & 0.944 & 0.943 & 0.001  & [-0.016, 0.019] \\
Global -- Detailed Cascade         & 0.944 & 0.959 & -0.015 & [-0.031, 0.0001] \\
Simple Full -- Simple Cascade      & 0.939 & 0.941 & -0.002 & [-0.018, 0.013] \\
Simple Full -- Detailed Full       & 0.939 & 0.943 & -0.004 & [-0.022, 0.016] \\
Simple Full -- Detailed Cascade    & 0.939 & 0.959 & -0.020 & [-0.037, -0.002] \\
Simple Cascade -- Detailed Full    & 0.941 & 0.943 & -0.002 & [-0.020, 0.017] \\
Simple Cascade -- Detailed Cascade & 0.941 & 0.959 & -0.018 & [-0.034, -0.003] \\
Detailed Full -- Detailed Cascade  & 0.943 & 0.959 & -0.016 & [-0.031, -0.003] \\
\addlinespace
\multicolumn{5}{l}{\textit{Opus 4.7}} \\
Global -- Simple Full              & 0.942 & 0.946 & -0.004 & [-0.014, 0.004] \\
Global -- Simple Cascade           & 0.942 & 0.942 & 0.000  & [-0.012, 0.012] \\
Global -- Detailed Full            & 0.942 & 0.955 & -0.014 & [-0.029, 0.003] \\
Global -- Detailed Cascade         & 0.942 & 0.950 & -0.008 & [-0.022, 0.007] \\
Simple Full -- Simple Cascade      & 0.946 & 0.942 & 0.004  & [-0.005, 0.014] \\
Simple Full -- Detailed Full       & 0.946 & 0.955 & -0.010 & [-0.025, 0.007] \\
Simple Full -- Detailed Cascade    & 0.946 & 0.950 & -0.004 & [-0.017, 0.011] \\
Simple Cascade -- Detailed Full    & 0.942 & 0.955 & -0.014 & [-0.030, 0.004] \\
Simple Cascade -- Detailed Cascade & 0.942 & 0.950 & -0.008 & [-0.021, 0.006] \\
Detailed Full -- Detailed Cascade  & 0.955 & 0.950 & 0.006  & [-0.008, 0.019] \\

\midrule

\multicolumn{5}{l}{\textbf{Panel B: Approach Types}} \\
\midrule
\textbf{Approach Types} &
\textbf{Type 1} &
\textbf{Type 2} &
\textbf{Difference} &
\textbf{95\% CI} \\
\midrule

\multicolumn{5}{l}{\textit{GPT-5.4}} \\
Cascade -- Full document           & 0.950 & 0.941 & 0.009 & [-0.002, 0.020] \\
Detailed prompts -- Simple prompts & 0.951 & 0.940 & 0.011 & [-0.004, 0.024] \\
Interaction                        & - & - & 0.014 & [-0.005, 0.035] \\

\addlinespace
\multicolumn{5}{l}{\textit{Opus 4.7}} \\
Cascade -- Full document           & 0.946 & 0.950 & -0.005 & [-0.014, 0.004] \\
Detailed prompts -- Simple prompts & 0.952 & 0.944 & 0.009  & [-0.005, 0.021] \\
Interaction                        & - & - & -0.002 & [-0.017, 0.014] \\

\bottomrule
\end{tabular}

\begin{tablenotes}[flushleft]
\footnotesize
\item \textit{Notes.} Entries report macro-averaged F1 scores across the
18 variables with at least ten positive instances in the bylaws benchmark.
In Panel A, Difference is the F1 score for Approach 1 minus the F1 score for
Approach 2. In Panel B, Cascade averages across Detailed Cascade and Simple Cascade, while Full Document averages across Detailed Full and Simple Full. Detailed Prompts averages across Detailed Full and Detailed Cascade, while Simple Prompts averages across Simple Full and Simple Cascade. Difference is Type 1 minus Type 2. The interaction measures whether the effect
of excerpting differs between Detailed and Simple prompts and is computed as ( Detailed Cascade $-$ Detailed Full ) $-$ ( Simple Cascade $-$ Simple Full ). Confidence intervals are percentile-based 95\% confidence intervals from document-level bootstraps with $R=10,000$ iterations.
\end{tablenotes}

\end{threeparttable}
\end{table*}

Table \ref{tab:bootstrap_charters} reports the model comparisons underlying the results on model comparisons in the charter context described in Section 7. Its structure mirrors that of Table \ref{tab:bootstrap_bylaws}, but it evaluates the charters benchmark instead of the bylaws benchmark. 

\begin{table*}[h]
\centering
\caption{Bootstrap Comparisons of Model Performance--Charters}
\label{tab:bootstrap_charters}
\small

\begin{threeparttable}

\begin{tabular}{l c c c c}
\toprule
\multicolumn{5}{l}{\textbf{Panel A: Frontier vs. Efficiency Models}} \\
\midrule
\textbf{Model Comparison} &
\textbf{Model 1} &
\textbf{Model 2} &
\textbf{Difference} &
\textbf{95\% CI} \\
\midrule

\multicolumn{5}{l}{\textit{Global}} \\
GPT-5.4 -- GPT-5.4-mini            & 0.896 & 0.795 & 0.101 & [0.076, 0.126] \\
Opus 4.7 -- Sonnet 4.6            & 0.932 & 0.900 & 0.031 & [0.014, 0.050] \\
Gemini Pro 2.5 -- Gemini Flash 2.5 & 0.899 & 0.893 & 0.007 & [-0.012, 0.025] \\

\addlinespace
\multicolumn{5}{l}{\textit{Detailed Cascade}} \\
GPT-5.4 -- GPT-5.4-mini            & 0.891 & 0.871 & 0.020 & [0.0002, 0.040] \\
Opus 4.7 -- Sonnet 4.6            & 0.898 & 0.873 & 0.025 & [0.007, 0.047] \\
Gemini Pro 2.5 -- Gemini Flash 2.5 & 0.896 & 0.887 & 0.009 & [-0.004, 0.023] \\

\midrule

\multicolumn{5}{l}{\textbf{Panel B: Detailed Cascade vs. Global}} \\
\midrule
\textbf{Model} &
\textbf{Cascade} &
\textbf{Global} &
\textbf{Difference} &
\textbf{95\% CI} \\
\midrule

\multicolumn{5}{l}{\textit{Frontier}} \\
GPT-5.4        & 0.891 & 0.896 & -0.006 & [-0.035, 0.020] \\
Opus 4.7       & 0.898 & 0.932 & -0.034 & [-0.061, -0.011] \\
Gemini Pro 2.5 & 0.896 & 0.899 & -0.003 & [-0.030, 0.022] \\

\addlinespace
\multicolumn{5}{l}{\textit{Efficiency}} \\
GPT-5.4-mini     & 0.871 & 0.795 & 0.075  & [0.042, 0.106] \\
Sonnet 4.6       & 0.873 & 0.900 & -0.028 & [-0.061, 0.001] \\
Gemini Flash 2.5 & 0.887 & 0.893 & -0.005 & [-0.034, 0.020] \\

\addlinespace
\multicolumn{5}{l}{\textit{Difference-in-Differences}} \\
GPT-5.4 -- GPT-5.4-mini            & 0.020 & 0.101 & -0.081 & [-0.111, -0.048] \\
Opus 4.7 -- Sonnet 4.6            & 0.025 & 0.031 & -0.006 & [-0.027, 0.014] \\
Gemini Pro 2.5 -- Gemini Flash 2.5 & 0.009 & 0.007 & 0.002  & [-0.021, 0.026] \\

\midrule

\multicolumn{5}{l}{\textbf{Panel C: Frontier Model Comparisons}} \\
\midrule
\textbf{Model Comparison} &
\textbf{Model 1} &
\textbf{Model 2} &
\textbf{Difference} &
\textbf{95\% CI} \\
\midrule

\multicolumn{5}{l}{\textit{Global}} \\
GPT-5.4 -- Opus 4.7        & 0.896 & 0.932 & -0.035 & [-0.054, -0.018] \\
GPT-5.4 -- Gemini Pro 2.5  & 0.896 & 0.899 & -0.003 & [-0.017, 0.012] \\
Opus 4.7 -- Gemini Pro 2.5 & 0.932 & 0.899 & 0.033  & [0.014, 0.053] \\

\addlinespace
\multicolumn{5}{l}{\textit{Detailed Cascade}} \\
GPT-5.4 -- Opus 4.7        & 0.891 & 0.898 & -0.007 & [-0.022, 0.006] \\
GPT-5.4 -- Gemini Pro 2.5  & 0.891 & 0.896 & -0.006 & [-0.017, 0.005] \\
Opus 4.7 -- Gemini Pro 2.5 & 0.898 & 0.896 & 0.002  & [-0.011, 0.014] \\

\bottomrule
\end{tabular}

\begin{tablenotes}[flushleft]
\footnotesize
\item \textit{Notes.} Entries report macro-averaged F1 scores across the 22 variables with at least ten positive instances in the charters benchmark. In Panel A, Difference is the frontier-model F1 score minus the corresponding efficiency-model F1 score. In Panel B, Difference is the Detailed Cascade F1 score minus the Global F1 score. The Difference-in-Differences rows report, for each model family, the performance gap between frontier and efficiency models under both Detailed Cascade and Global; Diff-in-Diff is the Detailed Cascade gap minus the Global gap. In Panel C, Difference is the first model's F1 score minus the second model's F1 score. Confidence intervals are percentile-based 95\% confidence intervals from document-level bootstraps with $R=10,000$ iterations.
\end{tablenotes}

\end{threeparttable}
\end{table*}

The results partly mirror those for bylaws, but with several important differences. Under the Global approach, the frontier model outperforms the efficiency-oriented model only for GPT and Claude, while the Gemini comparison includes zero. Under Detailed Cascade, unlike for bylaws, the frontier-efficiency gap is statistically significant for GPT and Claude. 

These results suggest that Detailed Cascade benefits efficiency-oriented models less in the charter setting. This conclusion is also supported by the within-model comparisons in Panel B. GPT-5.4-mini improves substantially under Detailed Cascade, whereas Sonnet 4.6 and Gemini Flash 2.5 have lower point estimates under Detailed Cascade than they have under Global (although their confidence intervals include zero). In line with this finding, the difference-in-differences estimate excludes zero only for the GPT family.

Another notable difference for charters is that all three frontier models have lower F1 point estimates under Detailed Cascade than under Global. Only for Opus 4.7 does the confidence interval exclude zero, however. 

Finally, Panel C shows that the frontier models are less interchangeable under Global than in the bylaws benchmark: Opus 4.7 outperforms both GPT-5.4 and Gemini Pro 2.5. These differences disappear under Detailed Cascade. This is consistent with the earlier finding that Opus 4.7 performs particularly well under Global in this setting.

Table \ref{tab:bootstrap_bylaws_charters} reports the detailed results underlying the comparisons between bylaw and charter performance in Section 7. These comparisons shed light on the question whether the same governance variables are recovered equally well from both types of documents. I restrict the comparison to the 14 variables with at least ten positive instances in both benchmarks. In each bootstrap iteration, both benchmarks are resampled independently at the document level. Across both Global and Detailed Cascade and for all three frontier models, performance is higher on bylaws, with confidence intervals that consistently exclude zero.

\begin{table*}[htb]
\centering
\caption{Bootstrap Comparisons of Performance on Bylaws and Charters}
\label{tab:bootstrap_bylaws_charters}
\small

\begin{threeparttable}

\begin{tabular}{l c c c c}
\toprule
\textbf{Model} &
\textbf{Bylaws} &
\textbf{Charters} &
\textbf{Difference} &
\textbf{95\% CI} \\
\midrule

\multicolumn{5}{l}{\textit{Global}} \\
GPT-5.4        & 0.949 & 0.866 & 0.083 & [0.052, 0.114] \\
Opus 4.7       & 0.942 & 0.908 & 0.034 & [0.005, 0.067] \\
Gemini Pro 2.5 & 0.940 & 0.864 & 0.077 & [0.044, 0.113] \\

\addlinespace
\multicolumn{5}{l}{\textit{Detailed Cascade}} \\
GPT-5.4        & 0.960 & 0.888 & 0.072 & [0.040, 0.109] \\
Opus 4.7       & 0.956 & 0.880 & 0.076 & [0.043, 0.114] \\
Gemini Pro 2.5 & 0.949 & 0.881 & 0.068 & [0.035, 0.108] \\

\bottomrule
\end{tabular}

\begin{tablenotes}[flushleft]
\footnotesize
\item \textit{Notes.} Entries report macro-averaged F1 scores across the 14 variables with at least ten positive instances in both the bylaws and charters benchmarks. Difference is the average F1 score for bylaws minus the average F1 score for charters. Confidence intervals are percentile-based 95\% confidence intervals from two independent document-level bootstraps with $R=10,000$ iterations.
\end{tablenotes}

\end{threeparttable}
\end{table*}

\FloatBarrier

\clearpage
\section{Comparisons against the Adjudicated Benchmark}
\label{app:adjudicated}

This Appendix replicates the results in Sections 6 and 7 using the Adjudicated Benchmark to measure extraction performance. 

Figure \ref{fig:boxplot_flagship_adj} replicates Figure \ref{fig:boxplot_flagship} from the main text. The boxplots sit slightly higher here than they do in the latter figure. This is a consequence of the fact that, as noted before, performance estimates obtained from comparisons with the DECODEM Benchmarks are likely conservative, and the corrections made to obtain the Adjudicated Benchmark at least partly account for that bias.

\begin{figure*}[htb!]
\centering
\includegraphics[width=\textwidth]{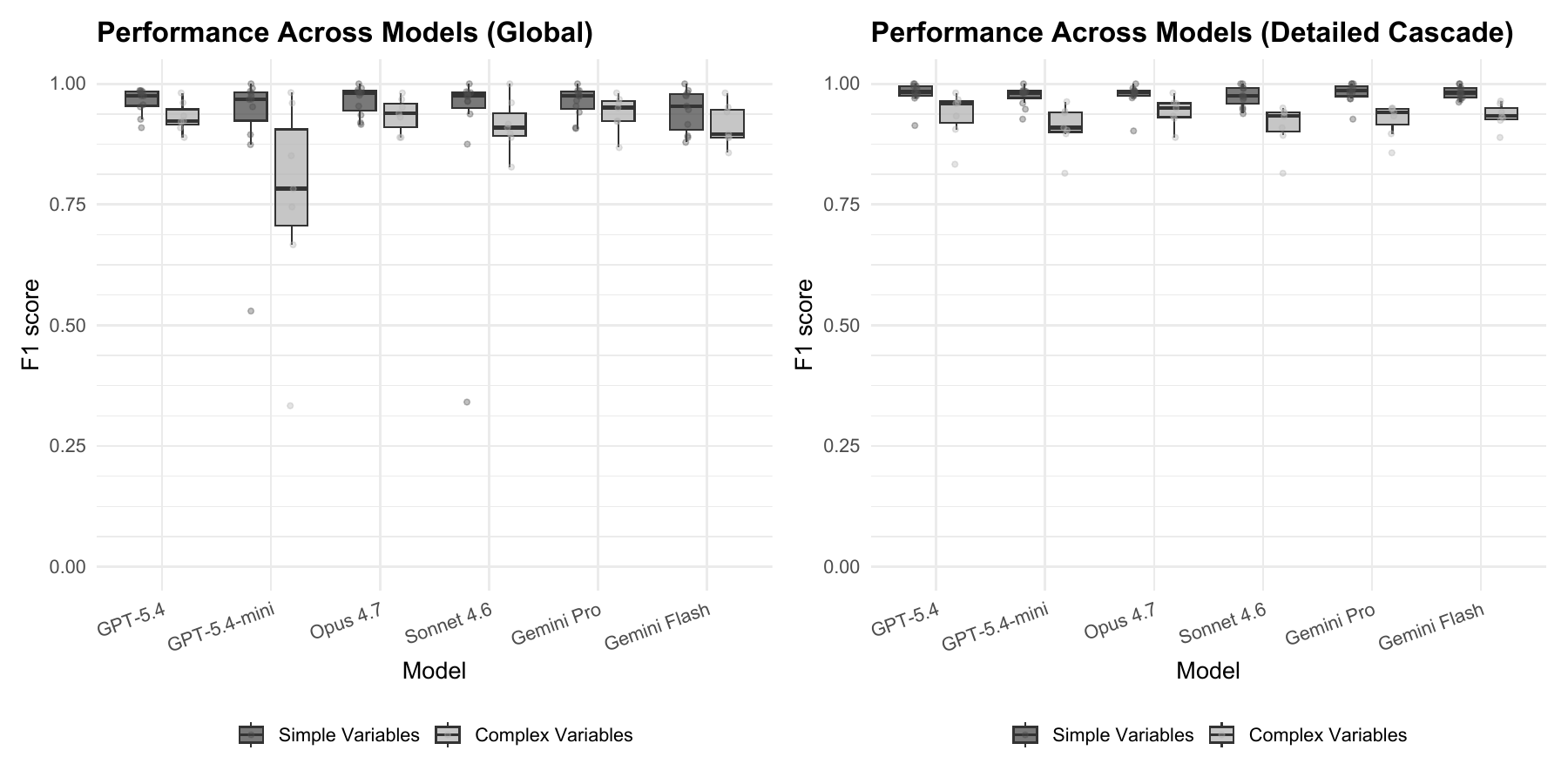}
\caption{Variable-level F1 scores for extraction of bylaw features across models (Adjudicated Benchmark).}
\label{fig:boxplot_flagship_adj}
\end{figure*}

Beyond this upward shift, the general pattern from the earlier figure is visible here as well. There is a notable gap between frontier models and efficiency models from each family under the Global approach. Under Detailed Cascade, this gap is much less pronounced. Performance also appears broadly similar across the three frontier models within each approach.

Figure \ref{fig:boxplot_highlevel_adj} replicates Figure \ref{fig:boxplot_highlevel}. As there, performance does not vary much across the different approaches. The three approaches not included in the analysis above also appear roughly comparable to Global. As with the previous results, Detailed Cascade appears to be the strongest overall performer here as well.

Table \ref{tab:bootstrap_bylaws_adj} reports results from the pairwise comparisons used in Section 6 to investigate the differences between models in more detail. The main patterns from the previous analysis replicate with the Adjudicated Benchmark. The three frontier models perform at similar levels, with differences even less pronounced than under the DECODEM Benchmark, where GPT-5.4 performed particularly well under Detailed Cascade. Frontier models gain less from the additional guidance that Detailed Cascade provides than the efficiency-oriented models, an effect that is again confirmed as systematic by the difference-in-differences estimates, whose confidence intervals uniformly exclude zero.

\begin{figure*}[htb!]
\centering
\includegraphics[width=\textwidth]{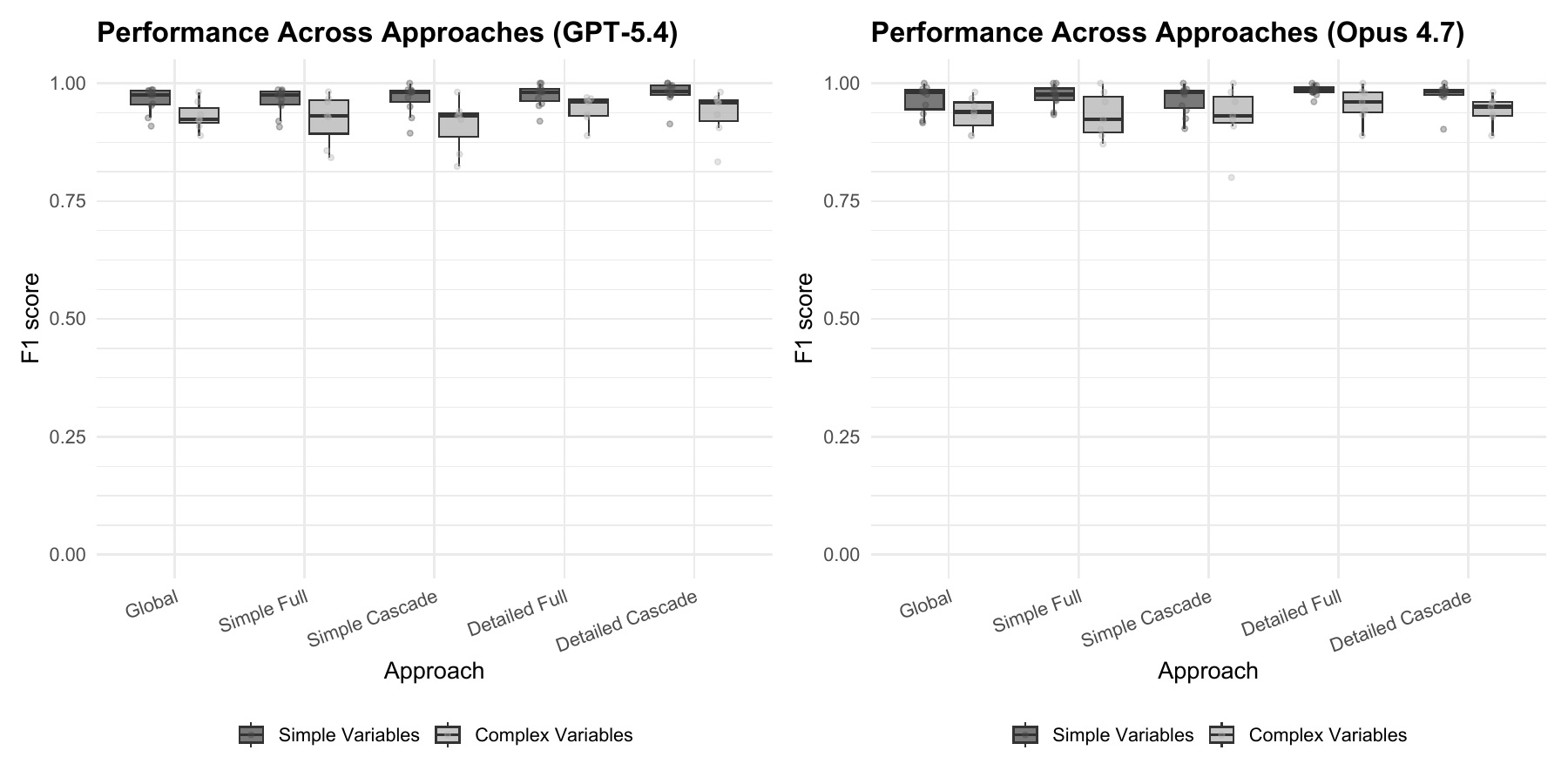}
\caption{Variable-level F1 scores for extraction of bylaw features across extraction approaches (Adjudicated Benchmark).}
\label{fig:boxplot_highlevel_adj}
\end{figure*}

\begin{figure*}[htb!]
\centering
\includegraphics[width=\textwidth]{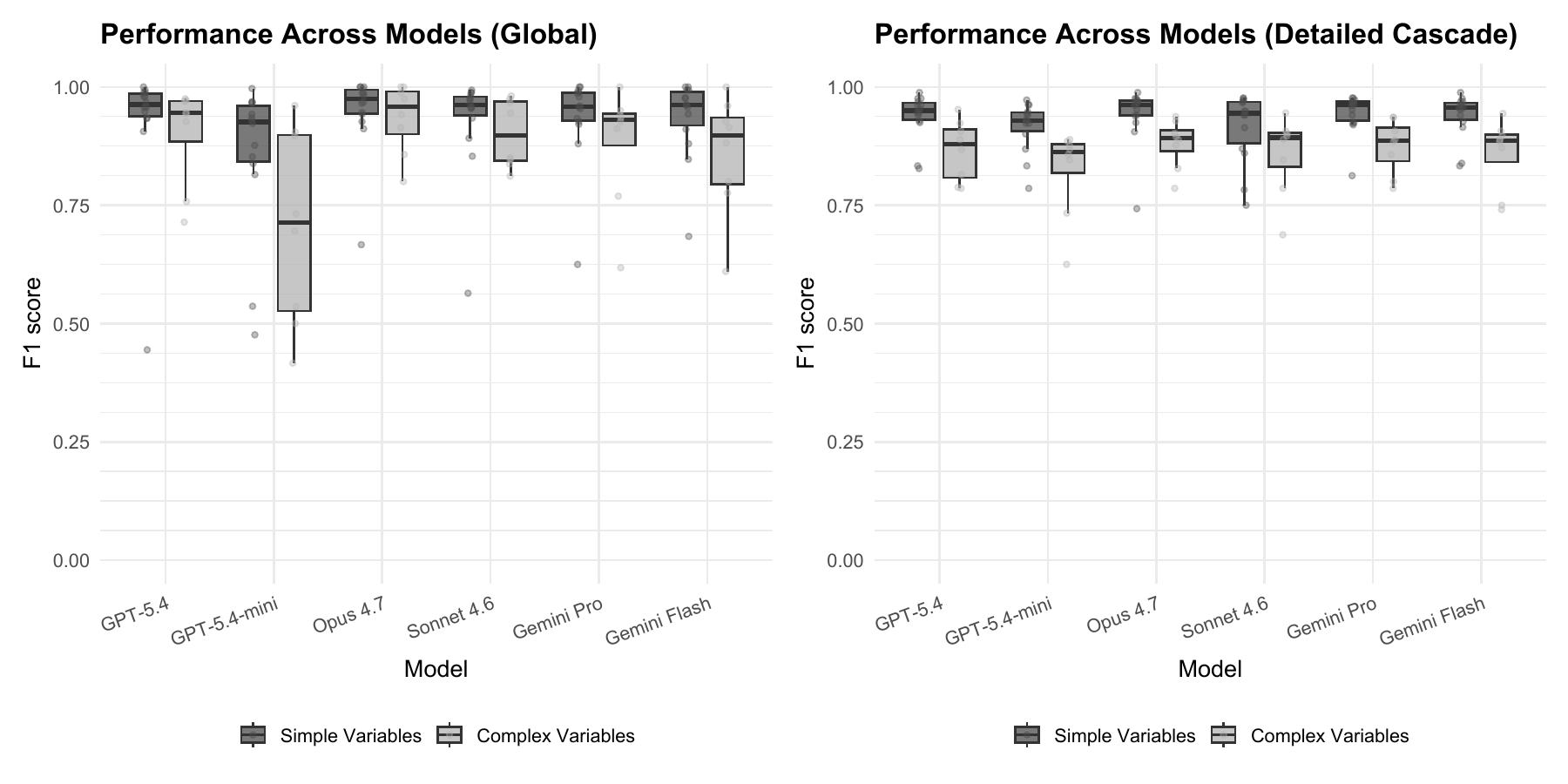}
\caption{Variable-level F1 scores for extraction of charter features across models (Adjudicated Benchmark).}
\label{fig:boxplot_flagship_charters_adj}
\end{figure*}

Finally, Figure \ref{fig:boxplot_flagship_charters_adj} replicates Figure \ref{fig:boxplot_flagship_charters}. As for the bylaws, substituting the Adjudicated Benchmark for the DECODEM Benchmark appears to marginally improve performance. As before, this is likely a function of the fact that the DECODEM Benchmark measures are conservative estimates of performance and that the Adjudicated Benchmark corrects for at least some of this downward bias. Beyond that, the results from the earlier figure appear mostly unchanged. Opus 4.7 under Global emerges as the strongest overall performer. Detailed Cascade does not generally improve performance relative to Global, and the only model that appears to gain substantially from Detailed Cascade is GPT-5.4-mini.

\begin{table*}[htb]
\centering
\caption{Bootstrap Comparisons of Model Performance--Bylaws (Adjudicated Benchmark)}
\label{tab:bootstrap_bylaws_adj}
\small

\begin{threeparttable}

\begin{tabular}{l c c c c}
\toprule
\multicolumn{5}{l}{\textbf{Panel A: Frontier vs. Efficiency Models}} \\
\midrule
\textbf{Model Comparison} &
\textbf{Model 1} &
\textbf{Model 2} & 
\textbf{Difference} &
\textbf{95\% CI} \\
\midrule

\multicolumn{5}{l}{\textit{Global}} \\
GPT-5.4 -- GPT-5.4-mini             & 0.952 & 0.858 & 0.094 & [0.072, 0.120] \\
Opus 4.7 -- Sonnet 4.6              & 0.954 & 0.910 & 0.044 & [0.030, 0.059] \\
Gemini Pro 2.5 -- Gemini Flash 2.5  & 0.953 & 0.933 & 0.020 & [0.007, 0.034] \\

\addlinespace
\multicolumn{5}{l}{\textit{Detailed Cascade}} \\
GPT-5.4 -- GPT-5.4-mini             & 0.962 & 0.949 & 0.013 & [0.003, 0.023] \\
Opus 4.7 -- Sonnet 4.6              & 0.963 & 0.949 & 0.014 & [0.000, 0.029] \\
Gemini Pro 2.5 -- Gemini Flash 2.5  & 0.959 & 0.963 & -0.004 & [-0.013, 0.004] \\

\midrule

\multicolumn{5}{l}{\textbf{Panel B: Detailed Cascade vs. Global}} \\
\midrule
\textbf{Model} &
\textbf{Cascade} &
\textbf{Global} &
\textbf{Difference} &
\textbf{95\% CI} \\
\midrule

\multicolumn{5}{l}{\textit{Frontier}} \\
GPT-5.4          & 0.962 & 0.952 & 0.010 & [-0.005, 0.025] \\
Opus 4.7         & 0.963 & 0.954 & 0.009 & [-0.006, 0.024] \\
Gemini Pro 2.5   & 0.959 & 0.953 & 0.006 & [-0.011, 0.023] \\

\addlinespace
\multicolumn{5}{l}{\textit{Efficiency}} \\
GPT-5.4-mini     & 0.949 & 0.858 & 0.092 & [0.070, 0.118] \\
Sonnet 4.6       & 0.949 & 0.910 & 0.040 & [0.022, 0.057] \\
Gemini Flash 2.5 & 0.963 & 0.933 & 0.030 & [0.013, 0.048] \\

\addlinespace
\multicolumn{5}{l}{\textit{Difference-in-Differences}} \\
GPT-5.4 -- GPT-5.4-mini             & 0.013 & 0.094 & -0.081 & [-0.109, -0.056] \\
Opus 4.7 -- Sonnet 4.6              & 0.014 & 0.044 & -0.031 & [-0.050, -0.011] \\
Gemini Pro 2.5 -- Gemini Flash 2.5  & -0.004 & 0.020 & -0.024 & [-0.040, -0.009] \\

\midrule

\multicolumn{5}{l}{\textbf{Panel C: Frontier Model Comparisons}} \\
\midrule
\textbf{Model Comparison} &
\textbf{Model 1} &
\textbf{Model 2} &
\textbf{Difference} &
\textbf{95\% CI} \\
\midrule

\multicolumn{5}{l}{\textit{Global}} \\
GPT-5.4 -- Opus 4.7                 & 0.952 & 0.954 & -0.002 & [-0.014, 0.009] \\
GPT-5.4 -- Gemini Pro 2.5           & 0.952 & 0.953 & -0.001 & [-0.013, 0.011] \\
Opus 4.7 -- Gemini Pro 2.5          & 0.954 & 0.953 & 0.001 & [-0.010, 0.013] \\

\addlinespace
\multicolumn{5}{l}{\textit{Detailed Cascade}} \\
GPT-5.4 -- Opus 4.7                 & 0.962 & 0.963 & -0.001 & [-0.017, 0.013] \\
GPT-5.4 -- Gemini Pro 2.5           & 0.962 & 0.959 & 0.003 & [-0.014, 0.018] \\
Opus 4.7 -- Gemini Pro 2.5          & 0.963 & 0.959 & 0.004 & [-0.008, 0.016] \\

\bottomrule
\end{tabular}

\begin{tablenotes}[flushleft]
\footnotesize
\item \textit{Notes.} Entries report macro-averaged F1 scores across the 18 variables with at least ten positive instances in the bylaws benchmark. In Panel A, Difference is the frontier-model F1 score minus the corresponding efficiency-model F1 score. In Panel B, Difference is the Detailed Cascade F1 score minus the Global F1 score. The Difference-in-Differences rows report, for each model family, the performance gap between frontier and efficiency models under both Detailed Cascade and Global; Diff-in-Diff is the Detailed Cascade gap minus the Global gap. In Panel C, Difference is the first model's F1 score minus the second model's F1 score. Confidence intervals are percentile-based 95\% confidence intervals from document-level bootstraps with $R=10,000$ iterations.
\end{tablenotes}

\end{threeparttable}
\end{table*}
\clearpage
\section{Error Overlap}
\label{app:errors}

This Appendix analyzes the overlap between the errors made by the various models and approaches. This analysis has potentially important implications for the ability to aggregate codings from different routines by way of ensemble coding\citep{frankenreiter2026measuring}.

Figure \ref{fig:model_errors} depicts how model errors are distributed across observations for each governance variable under the Global and Detailed Cascade approaches in the bylaws benchmark. The vertical axis indicates how many of the six models incorrectly classified a given observation. Observations are ordered from left to right by the number of errors.

\begin{figure*}[htb!]
\centering
\includegraphics[width=.8\textwidth]{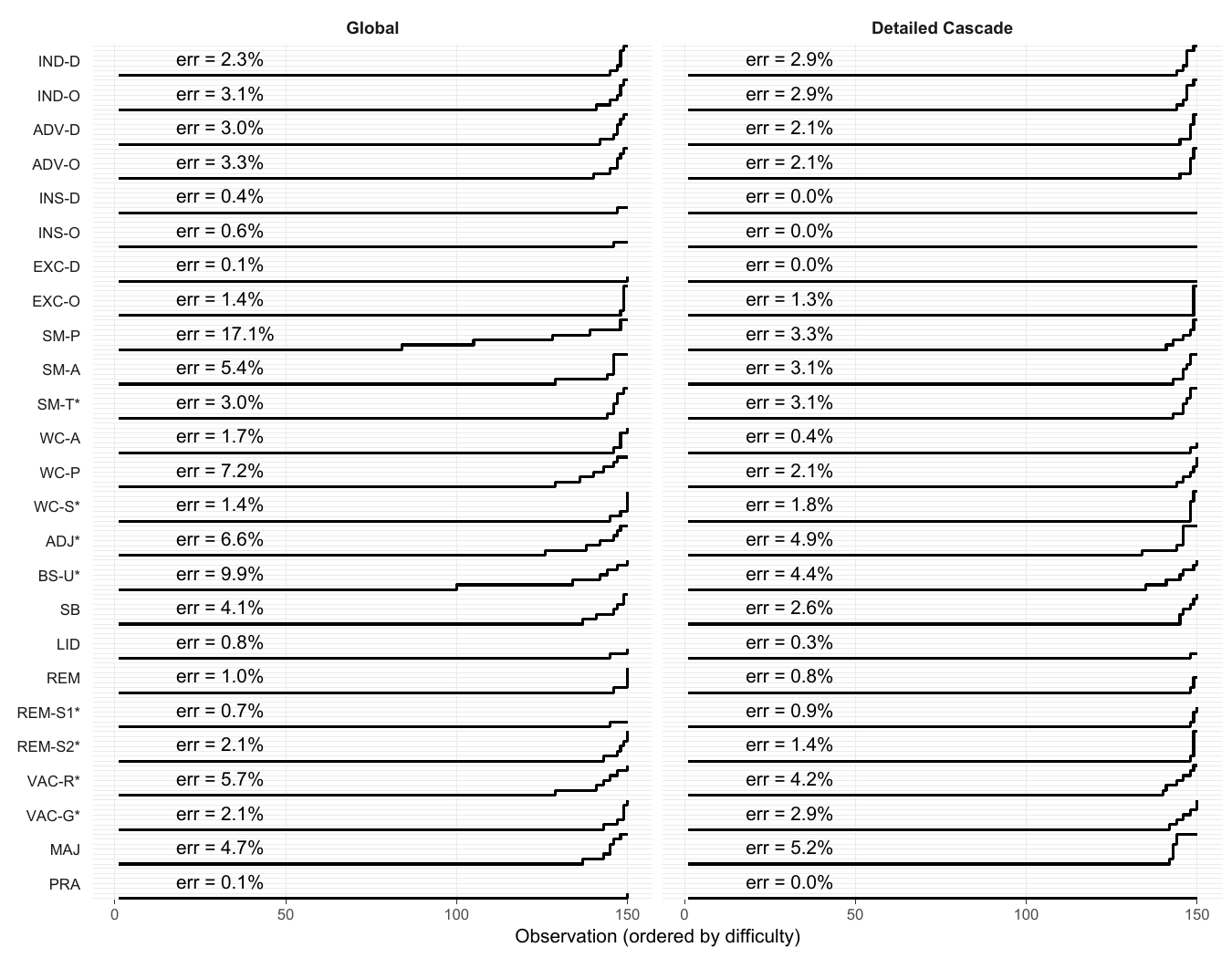}
\caption{Error concentration across models for the bylaws benchmark.}
\label{fig:model_errors}
\end{figure*}

Several patterns emerge. First, under both approaches, errors are concentrated among a relatively small subset of observations. By contrast, most observations are classified correctly by nearly all models, as evidenced by the long flat lines on the left side of each panel.

Second, the Global and Detailed Cascade approaches appear to show systematic differences. For Detailed Cascade, the curves usually stay close to zero for longer and rise sharply at the right end of the distribution. By contrast, the Global approach exhibits more gradual increases for many variables. This suggests that models using Detailed Cascade tend to fail on an overlapping set of difficult observations, while errors under Global are more dispersed across observations.

Figure \ref{fig:heatmap} explores this pattern further. The Figure reports pairwise Jaccard similarities (the proportion of shared errors among all observations misclassified by at least one of the two models) between the error sets of different models under each approach. High values indicate that models fail on the same observations.

\begin{figure*}[htb!]
\centering
\includegraphics[width=\textwidth]{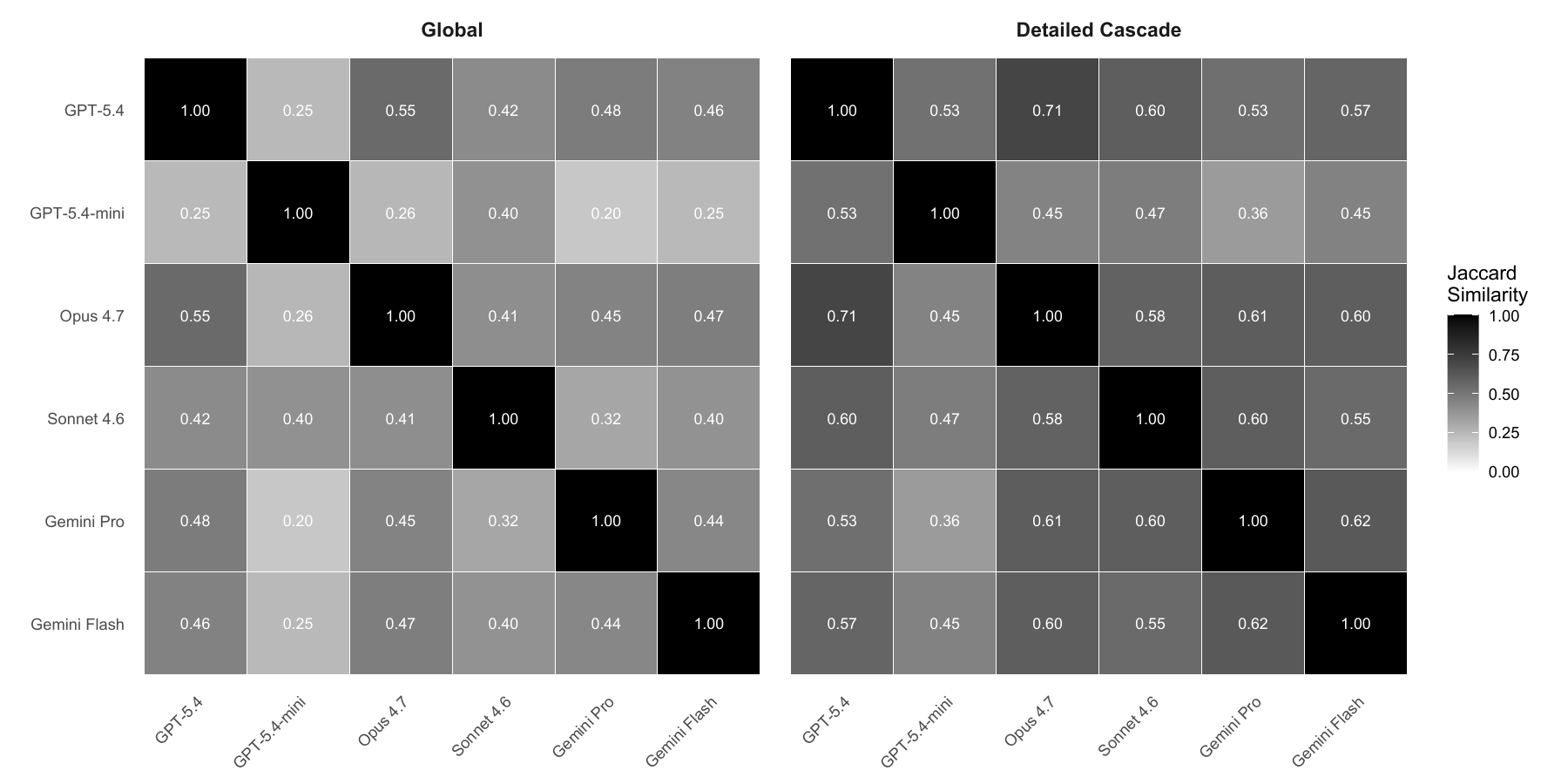}
\caption{Pairwise Jaccard similarities between model error sets for the bylaws benchmark.}
\label{fig:heatmap}
\end{figure*}

\begin{figure*}[b]
\centering
\includegraphics[width=.9\textwidth]{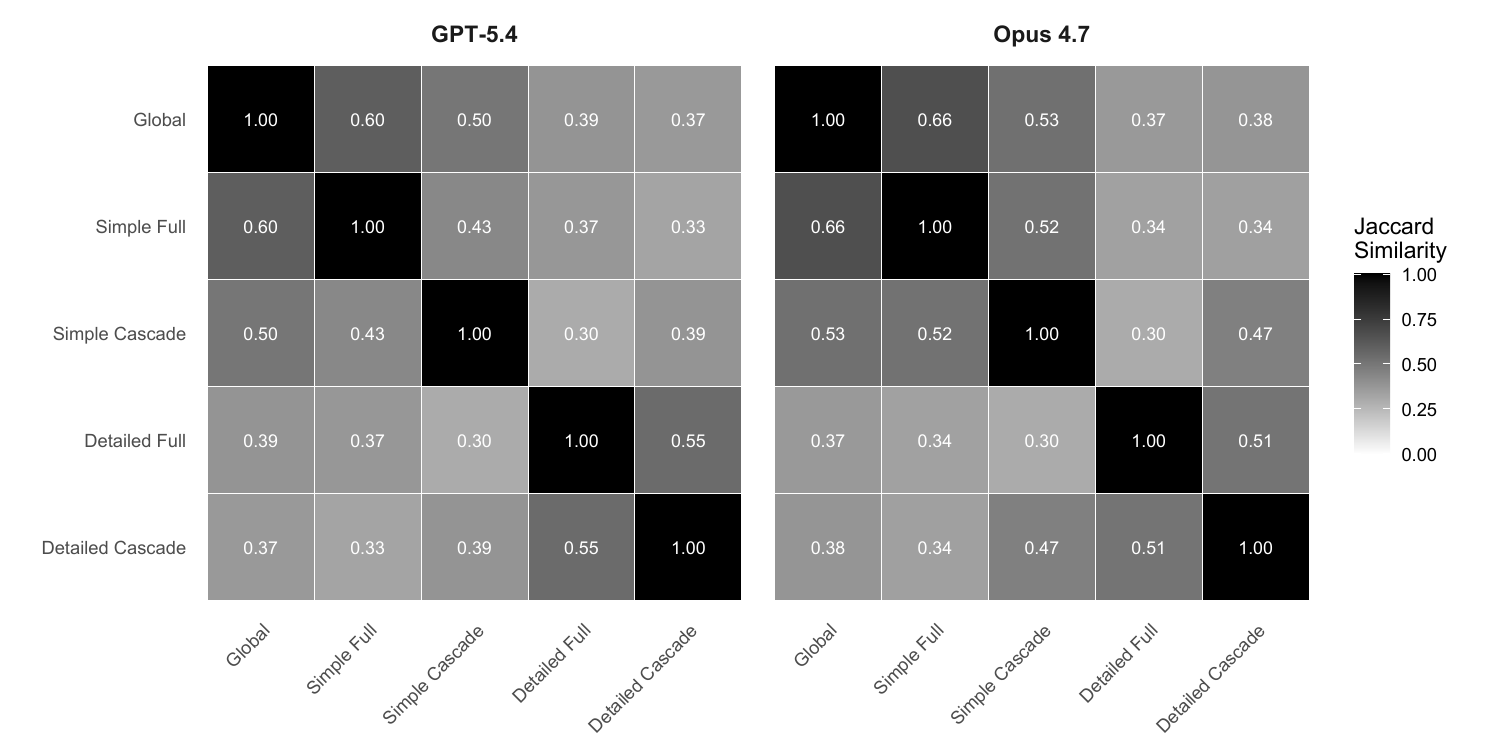}
\caption{Pairwise Jaccard similarities between approach error sets for the bylaws benchmark.}
\label{fig:heatmap_approaches}
\end{figure*}

The results corroborate the earlier finding of differences between the two extraction approaches. Under the Global approach, Jaccard similarities are substantially lower than under Detailed Cascade. For Global, Jaccard similarities generally range from 0.20 to 0.55, suggesting that models often make distinct mistakes when processing the full document context. Under Detailed Cascade, pairwise similarities range from 0.36 to 0.71. For every model pair, the Jaccard similarity under Detailed Cascade exceeds the corresponding value for Global. This suggests that models tend to make more similar mistakes under Detailed Cascade than they do under Global.

Figure \ref{fig:heatmap_approaches} repeats this analysis for the five extraction approaches, both under GPT-5.4 and under Opus 4.7. Extraction approaches that share common features, such as Detailed Cascade and Detailed Full, exhibit higher overlap, while extraction approaches that use different building blocks, such as Detailed Cascade and Simple Full, show less overlap. These results suggest that the choice of extraction architecture affects not only overall performance levels but also which observations are misclassified.

Figure \ref{fig:heatmap_charters} replicates Figure \ref{fig:heatmap} for the charters benchmark. The overall pattern is similar. Errors show relatively high overlap under Detailed Cascade and less overlap under Global. If anything, the differences are even more pronounced here than in the bylaws setting. Importantly, this suggests that the differences between Global and Detailed Cascade observed for bylaws are not merely an artifact of Detailed Cascade performing better overall.

\begin{figure*}[thb]
\centering
\includegraphics[width=\textwidth]{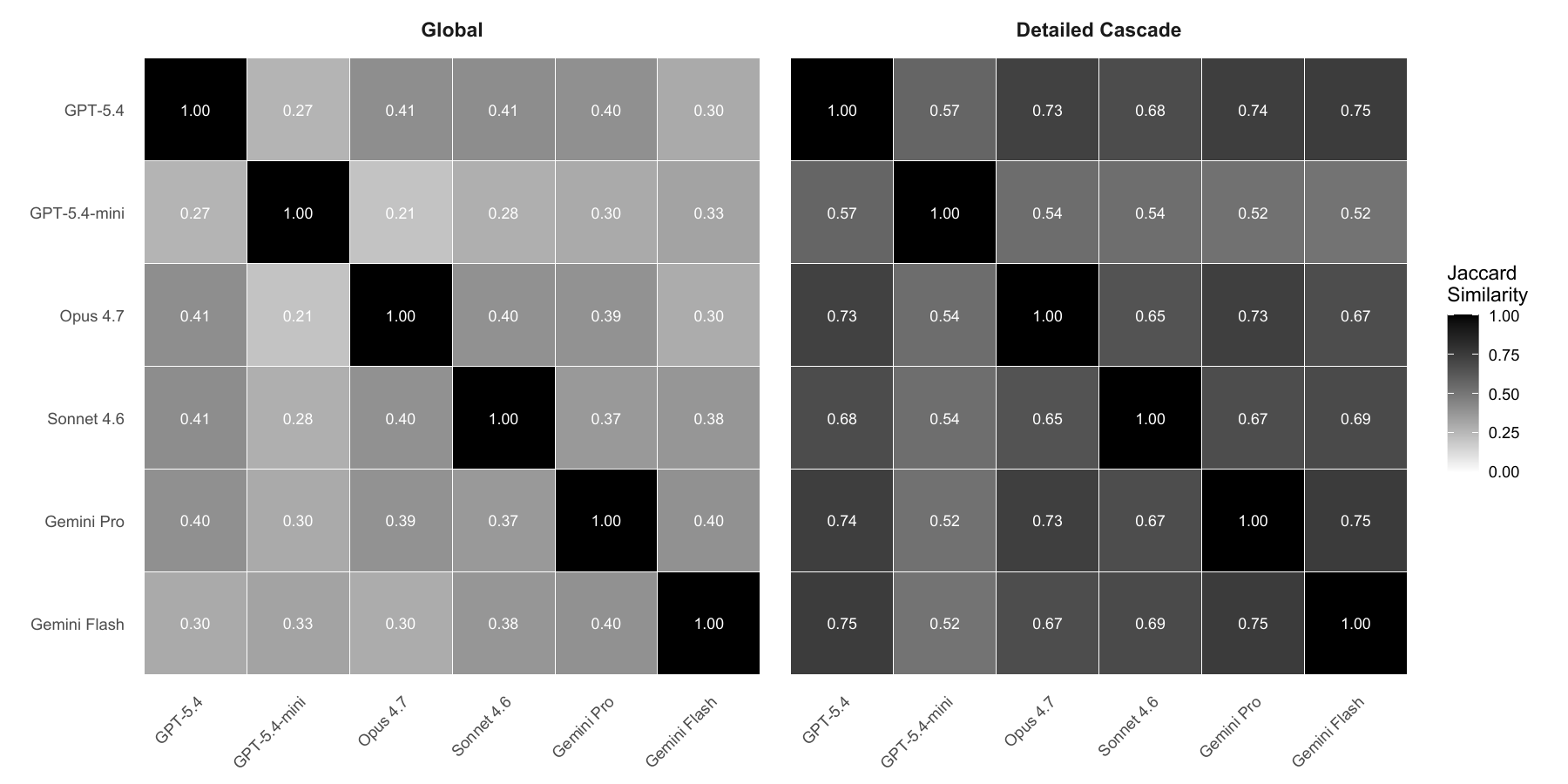}
\caption{Pairwise Jaccard similarities between model error sets for the charters benchmark.}
\label{fig:heatmap_charters}
\end{figure*}

Taken together, the findings suggest that Detailed Cascade induces substantially more correlated failure patterns across models than Global. One possible interpretation is that Detailed Cascade's more hands-on features create shared sources of error. This finding is supported by the comparison between approaches, which suggest that common features such as prompt design and the excerpting step may contribute to this pattern. By contrast, the Global approach appears to leave more room for model-specific differences in document processing and legal interpretation, resulting in less correlated error patterns.

\clearpage
\section{Variable-Level Results}
\label{app:variable-level}

This Appendix investigates performance of the various extraction approaches on the bylaws benchmark at the variable level.

Table \ref{tab:variable_level_results} reports results for GPT-5.4 and Opus 4.7. Several patterns emerge. First, a large set of variables---including indemnification and advancement provisions---can be extracted with very high accuracy across approaches and models. For these variables, F1 scores are uniformly close to one, and differences across approaches and models are negligible. A similar pattern holds for a number of less frequent variables, such as insurance, proxy access and lead independent director provisions, which also achieve near-perfect performance despite the small number of positive instances. What these variables share is that they tend to be expressed in relatively standardized language across bylaws, with little meaningful variation that would affect the ultimate coding decision. That said, indemnification, advancement, and insurance provisions vary in whether they require or merely allow the company to reimburse or provide insurance for the respective actor, and the models appear to correctly distinguish between the two types of provisions.

\begin{table*}[htb!]
\centering
\caption{Variable-Level Performance for GPT-5.4 and Opus 4.7 on the Bylaws Benchmark}
\label{tab:variable_level_results}
\footnotesize

\begin{threeparttable}

\begin{tabular}{l r c c c c c c c c c c c}
\toprule
\textbf{Variable} & \textbf{Count}
& \textbf{Global} & \textbf{Simple} & \textbf{Simple}
& \textbf{Detailed} & \textbf{Detailed}
&
& \textbf{Global} & \textbf{Simple} & \textbf{Simple}
& \textbf{Detailed} & \textbf{Detailed} \\
&
& & \textbf{Full} & \textbf{Cascade}
& \textbf{Full} & \textbf{Cascade}
&
& & \textbf{Full} & \textbf{Cascade}
& \textbf{Full} & \textbf{Cascade} \\
\midrule
& & \multicolumn{5}{c}{\textbf{GPT-5.4}}
& & \multicolumn{5}{c}{\textbf{Opus 4.7}} \\
\cmidrule{3-7} \cmidrule{9-13}

\multicolumn{7}{l}{\textit{Exculpation / Indemnification}} \\
IND-D & 108
& \textbf{0.986} & \textbf{0.986} & \textbf{0.986} & 0.981 & 0.977
& & 0.981 & 0.981 & \textbf{0.986} & \textbf{0.986} & 0.977 \\
IND-O & 107
& \textbf{0.986} & \textbf{0.986} & 0.986 & 0.986 & 0.977
& & 0.967 & 0.967 & 0.976 & \textbf{0.986} & 0.977 \\
ADV-D & 79
& 0.981 & 0.981 & \textbf{0.988} & \textbf{0.988} & 0.981
& & 0.969 & 0.969 & 0.981 & 0.975 & \textbf{0.981} \\
ADV-O & 79
& 0.981 & 0.981 & \textbf{0.988} & \textbf{0.988} & 0.981
& & 0.962 & 0.962 & 0.975 & 0.975 & \textbf{0.981} \\
INS-D & 6
& 6/0/0 & 6/0/0 & 6/0/0 & 6/0/0 & 6/0/0
& & 6/0/0 & 6/0/0 & 6/0/0 & 6/0/0 & 6/0/0 \\
INS-O & 6
& 6/0/0 & 6/0/0 & 6/0/0 & 6/0/0 & 6/0/0
& & 6/0/0 & 6/0/0 & 6/0/0 & 6/0/0 & 6/0/0 \\
EXC-D & 3
& 3/0/0 & 3/1/0 & 3/3/0 & 3/0/0 & 3/0/0
& & 3/0/0 & 3/0/0 & 3/0/0 & 3/0/0 & 3/0/0 \\
EXC-O & 3
& 1/0/2 & 1/1/2 & 1/3/2 & 1/0/2 & 1/0/2
& & 1/0/2 & 1/0/2 & 1/0/2 & 1/0/2 & 1/0/2 \\

\addlinespace
\multicolumn{7}{l}{\textit{Shareholder Meetings}} \\
SM-P & 75
& 0.913 & 0.966 & 0.913 & 0.973 & \textbf{0.980}
& & 0.936 & 0.966 & 0.905 & \textbf{0.987} & 0.980 \\
SM-A & 50
& 0.952 & 0.952 & 0.952 & 0.941 & \textbf{0.962}
& & 0.952 & 0.962 & 0.952 & \textbf{0.970} & 0.962 \\
SM-T* & 49
& 0.950 & 0.950 & 0.949 & 0.949 & \textbf{0.970}
& & 0.961 & 0.950 & 0.950 & 0.970 & \textbf{0.970} \\
WC-P & 32
& \textbf{0.985} & 0.952 & 0.970 & 0.968 & 0.980
& & 0.954 & 0.938 & 0.952 & \textbf{0.984} & \textbf{0.984} \\
WC-A & 50
& 0.942 & 0.916 & 0.940 & 0.961 & \textbf{1.000}
& & 0.906 & 0.923 & 0.932 & \textbf{0.970} & 0.961 \\
WC-S* & 5
& 4/0/1 & 3/1/2 & 1/0/4 & 4/0/1 & 3/0/2
& & 4/0/1 & 4/1/1 & 3/0/2 & 3/0/2 & 2/0/3 \\
ADJ* & 49
& 0.913 & 0.852 & 0.922 & \textbf{0.960} & 0.949
& & 0.929 & 0.913 & 0.922 & \textbf{0.970} & 0.949 \\

\addlinespace
\multicolumn{7}{l}{\textit{Board Structure}} \\
BS-U* & 60
& 0.922 & 0.912 & 0.865 & 0.950 & \textbf{0.967}
& & 0.930 & 0.869 & 0.908 & 0.958 & \textbf{0.966} \\
SB & 40
& 0.909 & 0.920 & 0.918 & 0.920 & \textbf{0.938}
& & 0.920 & \textbf{0.988} & 0.950 & 0.951 & 0.927 \\
LID & 8
& 8/1/0 & 8/1/0 & 8/0/0 & 8/0/0 & 8/0/0
& & 8/0/0 & 8/0/0 & 7/0/1 & 8/0/0 & 8/0/0 \\

\addlinespace
\multicolumn{7}{l}{\textit{Director Elections}} \\
REM* & 26
& \textbf{1.000} & 0.963 & \textbf{1.000} & 0.981 & \textbf{1.000}
& & 0.962 & 0.962 & \textbf{1.000} & 0.981 & \textbf{1.000} \\
REM-S1* & 16
& 0.933 & \textbf{0.967} & 0.933 & 0.933 & 0.933
& & 0.968 & \textbf{1.000} & \textbf{1.000} & 0.933 & 0.933 \\
REM-S2* & 7
& 7/0/0 & 7/0/0 & 7/0/0 & 5/0/2 & 5/0/2
& & 7/0/0 & 7/0/0 & 7/0/0 & 6/0/1 & 7/0/0 \\
VAC-R* & 25
& 0.885 & 0.846 & \textbf{0.917} & 0.800 & 0.909
& & 0.863 & 0.863 & 0.766 & \textbf{0.889} & 0.868 \\
VAC-G* & 11
& 0.880 & \textbf{0.906} & 0.833 & 0.830 & 0.880
& & 0.880 & \textbf{0.917} & \textbf{0.917} & 0.800 & 0.800 \\
MAJ & 35
& \textbf{0.909} & \textbf{0.909} & 0.875 & \textbf{0.909} & 0.875
& & \textbf{0.909} & 0.892 & 0.875 & \textbf{0.909} & 0.875 \\
PRA & 11
& 0.957 & 0.957 & \textbf{1.000} & 0.957 & \textbf{1.000}
& & 1.000 & 1.000 & 1.000 & 1.000 & 1.000 \\

\bottomrule
\end{tabular}

\begin{tablenotes}[flushleft]
\footnotesize
\item \textit{Notes.} Entries report F1 scores by variable and extraction
approach for variables with Count $\geq 10$ and TP/FP/FN (true positives /
false positives / false negatives) for other variables. Count indicates the number of positive instances for the respective variable in the benchmark dataset.
Complex Variables are marked with a *.
\end{tablenotes}

\end{threeparttable}
\end{table*}

Second, for a number of variables, performance remains high but exhibits more variation across approaches. This is particularly visible for the shareholder meeting and board structure provisions. In these cases, F1 scores remain close to one, but differences across approaches are more pronounced than for the near-uniform variables discussed above. Across these variables, the two approaches using detailed prompts appear to emerge as the stronger approaches. For GPT-5.4 in particular, Detailed Cascade appears to be the strongest overall performer for this category of variables. This mirrors the finding from Section 6 that Detailed Cascade performs somewhat better than several rival approaches on the bylaws benchmark. However, the results here indicate that any advantages are concentrated within certain families of variables.

Third, the Complex Variables exhibit substantially greater variation across approaches and, on average, lower performance than the Simple Variables. Consider ADJ as an example. Here, performance differs markedly across approaches even when using the same simple prompting questions, while approaches based on more detailed prompting perform better. Conversely, for REM-S2, the simpler approaches outperform the detailed ones. These patterns illustrate that performance differences are not driven by a single superior approach, but instead depend on the structure of the underlying provision and, potentially, on specific features of the prompts used.

Fourth, performance is sometimes lower for relatively rare provisions, both in terms of their frequency in the dataset and their prevalence in practice. The comparison between EXC-D and EXC-O is instructive. These provisions often appear in the same bylaw documents, yet models are consistently better at detecting the former than the latter. One possible explanation is that officer-exculpating waivers were impermissible in most jurisdictions until very recently, while director-protecting waivers were standard in many charters\cite{frankenreiter2026sticky_charters}. As a result, models may have internalized the expectation that exculpation provisions primarily protect directors and may fail to identify analogous provisions for officers even when they appear in bylaws.

Taken together, the results reinforce the results from Figure \ref{fig:boxplot_highlevel}. Most governance provisions can be identified with high accuracy irrespective of the extraction approach. A smaller set of variables accounts for most of the observed variation in performance between approaches. Overall, differences across extraction approaches are modest, and while Detailed Cascade appears to perform best in some settings, no single method consistently dominates. 
\clearpage
\section{Repeated Runs}
\label{app:repeated-runs}

This Appendix reports results for repeated model runs. The purpose of this exercise is to assess whether the results are sensitive to run-to-run variation in model outputs.

To check for run-to-run variation, I re-ran the extraction twice for all models under the Global approach on the bylaws benchmark. I also repeated the Detailed Cascade approach for GPT-5.4 and Opus 4.7 on the bylaws benchmark and the Global approach for the same two models on the charters benchmark. Together with the original extraction run, this yields three sets of predictions for each model-approach combination reported here.

Table \ref{tab:rerun_disagreement} first examines the amount of variation at the level of individual predicted labels. For the frontier models, only very limited variation can be observed. For GPT-5.4 and Opus 4.7, the average pairwise disagreement rate is around half a percent under both approaches on the bylaws benchmark and under Global for the charters benchmark. Gemini Pro 2.5 exhibits somewhat more variation, with an average disagreement rate of approximately 1.1\% under Global for the bylaws benchmark.

\begin{table*}[htb]
\centering
\caption{Prediction-Level Variation Across Repeated Runs}
\label{tab:rerun_disagreement}
\small

\begin{threeparttable}

\begin{tabular}{l ccccc}
\toprule
\textbf{Model} &
\textbf{Observations} &
\shortstack{\textbf{Run 2 vs.}\\\textbf{Run 1}} &
\shortstack{\textbf{Run 3 vs.}\\\textbf{Run 1}} &
\shortstack{\textbf{Run 3 vs.}\\\textbf{Run 2}} &
\shortstack{\textbf{Average Dis-}\\\textbf{agreement (\%)}} \\
\midrule

\multicolumn{6}{l}{\textbf{Panel A: Bylaws--Global}} \\
\midrule

GPT-5.4
& 3,750 & 18 & 23 & 17 & 0.52 \\

GPT-5.4-mini
& 3,750 & 88 & 79 & 85 & 2.24 \\

Opus 4.7
& 3,750 & 18 & 24 & 22 & 0.57 \\

Sonnet 4.6
& 3,750 & 36 & 38 & 30 & 0.92 \\

Gemini Pro 2.5
& 3,750 & 51 & 38 & 29 & 1.05 \\

Gemini Flash 2.5
& 3,750 & 87 & 90 & 27 & 1.81 \\

\midrule

\multicolumn{6}{l}{\textbf{Panel B: Bylaws--Detailed Cascade}} \\
\midrule

GPT-5.4
& 3,750 & 13 & 17 & 18 & 0.43 \\

Opus 4.7
& 3,750 & 14 & 14 & 8 & 0.32 \\

\midrule

\multicolumn{6}{l}{\textbf{Panel C: Charters--Global}} \\
\midrule

GPT-5.4
& 7,800 & 48 & 58 & 28 & 0.57 \\

Opus 4.7
& 7,800 & 39 & 42 & 31 & 0.48 \\

\bottomrule
\end{tabular}

\begin{tablenotes}[flushleft]
\footnotesize
\item \textit{Notes.} The three ``Runs'' columns report the number of predicted labels that differ between the respective runs. Average Disagreement reports the mean pairwise disagreement rate for the three run-to-run comparisons.
\end{tablenotes}

\end{threeparttable}
\end{table*}

The efficiency-oriented models also show more variation. The average disagreement rate under Global ranges from 0.9\% for Sonnet 4.6 to 2.2\% for GPT-5.4-mini. Even for these models, however, predictions are identical across repeated runs for the large majority of observations.

Figure \ref{fig:heatmap_variation} displays variation at the variable level. Each cell reports the average disagreement rate between two runs across the three runs for a given variable and extraction routine / model combination. Consistent with the aggregate results, disagreement is low for most variables, with many cells indicating that all predictions are identical across the three runs.

\begin{figure*}[htb!]
\centering
\includegraphics[width=\textwidth]{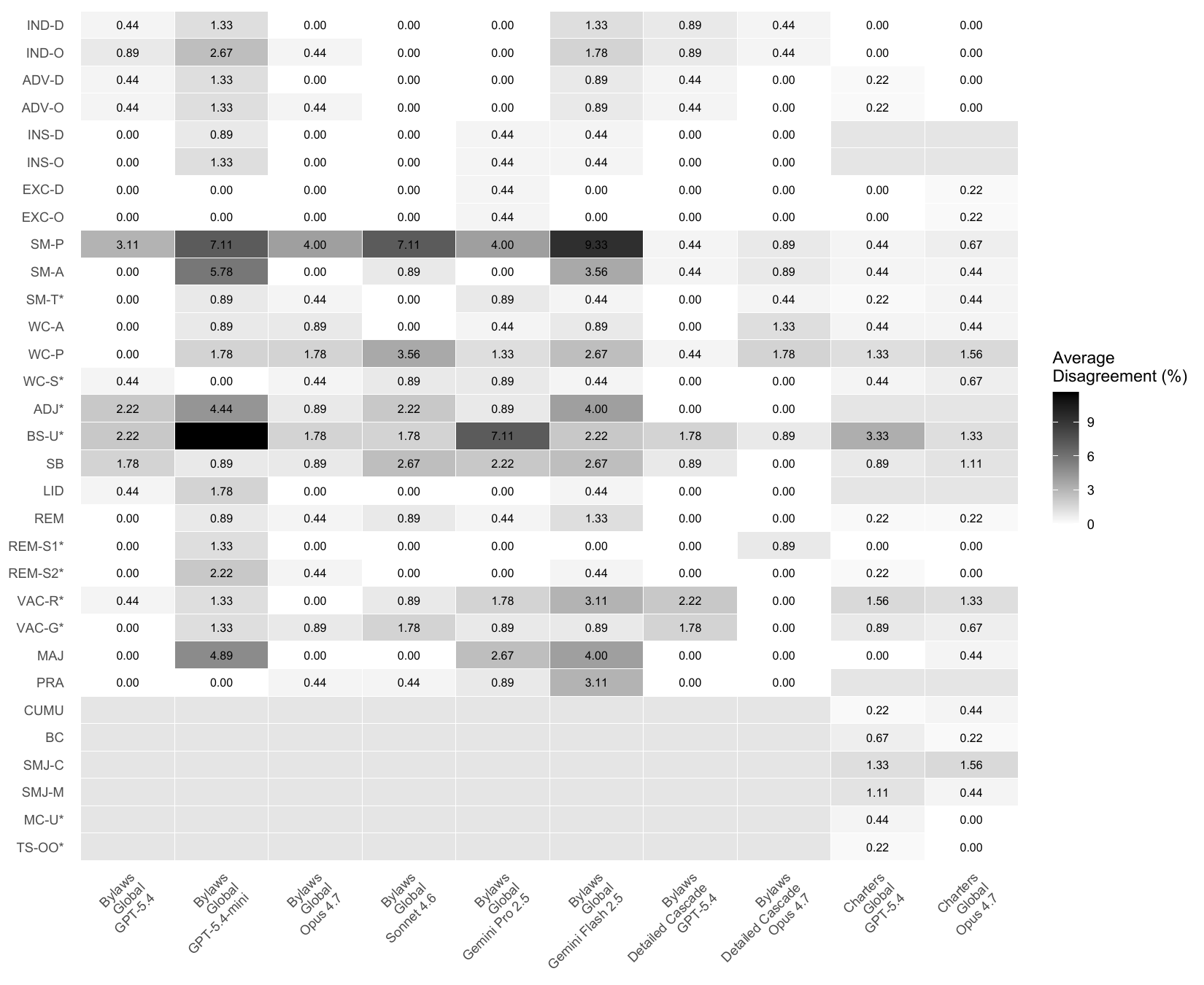}
\caption{Prediction-level rerun variation by variable.}
\label{fig:heatmap_variation}
\end{figure*}

Instead, variation is concentrated in a relatively small number of variables. One example is SM-P, which exhibits comparatively high disagreement rates across several models under the Global approach. The pattern also suggests a possible relationship between variable-level run-to-run variation and performance as reported in Section \ref{app:variable-level} of the Appendix. Variables that exhibit greater variation here also tend to exhibit lower performance there.

Table \ref{tab:repeated_runs} examines how the variation at the level of individual variables translates into variation in extraction performance. The table reports macro-averaged F1 scores for each run, computed over all variables with at least 10 positive instances in the respective benchmark. 

\begin{table*}[htb]
\centering
\caption{Performance Across Repeated Runs}
\label{tab:repeated_runs}
\small

\begin{threeparttable}

\begin{tabular}{l l c c c}
\toprule
\textbf{Model} &
\textbf{Benchmark} &
\textbf{Run 1} &
\textbf{Run 2} &
\textbf{Run 3}\\
\midrule

\multicolumn{5}{l}{\textbf{Panel A: Bylaws--Global}} \\
\midrule

GPT-5.4
  & DECODEM      & 0.944 & 0.944 & 0.946  \\
  & Adjudicated  & 0.952 & 0.953 & 0.954  \\

\addlinespace
GPT-5.4-mini
  & DECODEM      & 0.857 & 0.843 & 0.847  \\
  & Adjudicated  & 0.858 & 0.843 & 0.850  \\

\addlinespace
Opus 4.7
  & DECODEM      & 0.942 & 0.940 & 0.939 \\
  & Adjudicated  & 0.954 & 0.952 & 0.954  \\

\addlinespace
Sonnet 4.6
  & DECODEM      & 0.897 & 0.893 & 0.898  \\
  & Adjudicated  & 0.910 & 0.905 & 0.908 \\

\addlinespace
Gemini Pro 2.5
  & DECODEM      & 0.944 & 0.941 & 0.941 \\
  & Adjudicated  & 0.953 & 0.952 & 0.954  \\

\addlinespace
Gemini Flash 2.5
  & DECODEM      & 0.927 & 0.925 & 0.922 \\
  & Adjudicated  & 0.933 & 0.924 & 0.919 \\

\midrule

\multicolumn{5}{l}{\textbf{Panel B: Bylaws--Detailed Cascade}} \\
\midrule

GPT-5.4
  & DECODEM      & 0.959 & 0.954 & 0.956  \\
  & Adjudicated  & 0.962 & 0.963 & 0.960  \\

\addlinespace
Opus 4.7
  & DECODEM      & 0.950 & 0.948 & 0.949  \\
  & Adjudicated  & 0.963 & 0.962 & 0.963  \\

\midrule

\multicolumn{5}{l}{\textbf{Panel C: Charters--Global}} \\
\midrule

GPT-5.4
  & DECODEM      & 0.896 & 0.905 & 0.902  \\
  & Adjudicated  & 0.917 & 0.919 & 0.917  \\

\addlinespace
Opus 4.7
  & DECODEM      & 0.932 & 0.934 & 0.933  \\
  & Adjudicated  & 0.943 & 0.946 & 0.946  \\

\bottomrule
\end{tabular}

\begin{tablenotes}[flushleft]
\footnotesize
\item \textit{Notes.} Entries report macro-averaged F1 scores across variables
with at least ten positive instances in the DECODEM Benchmark. This includes
18 variables for the bylaws benchmark and 22 variables for the charters
benchmark. Run 1 is the original extraction run used in the main analysis.
\end{tablenotes}

\end{threeparttable}
\end{table*}

While the results for frontier models are again highly stable, efficiency-oriented models show more variation. For GPT-5.4-mini, the range is approximately 1.4--1.5 F1 points, depending on the benchmark used for evaluation. Gemini Flash 2.5 exhibits similar variation under the Adjudicated Benchmark. Yet even these differences are relatively small compared to the performance differences described in the main text and do not alter any of the paper's substantive conclusions.

Taken together, the repeated-run results suggest that the findings reported in the paper are not sensitive to a particular stochastic realization of the models. Run-to-run variation is particularly small for the frontier models, while the efficiency-oriented models exhibit somewhat greater variability. Both individual predictions and aggregate performance nevertheless remain highly stable across runs.
\clearpage
\section{Document Length and Extraction Performance}
\label{app:doclength}

This Appendix examines whether extraction performance deteriorates as the length of the underlying document increases. The analysis focuses on the Global approach, for which the model receives the full document as input, and on the three frontier models.

Figure \ref{fig:errordistribution2} plots the number of extraction errors against document length for bylaws, distinguishing between full restatements and partial amendments. Partial amendments are generally much shorter than full restatements and also exhibit substantially fewer extraction errors. This difference is not surprising. Partial amendments typically address only a small subset of governance provisions. Therefore, their contents are easier to extract.

\begin{figure*}[htb!]
\centering
\includegraphics[width=.7\textwidth]{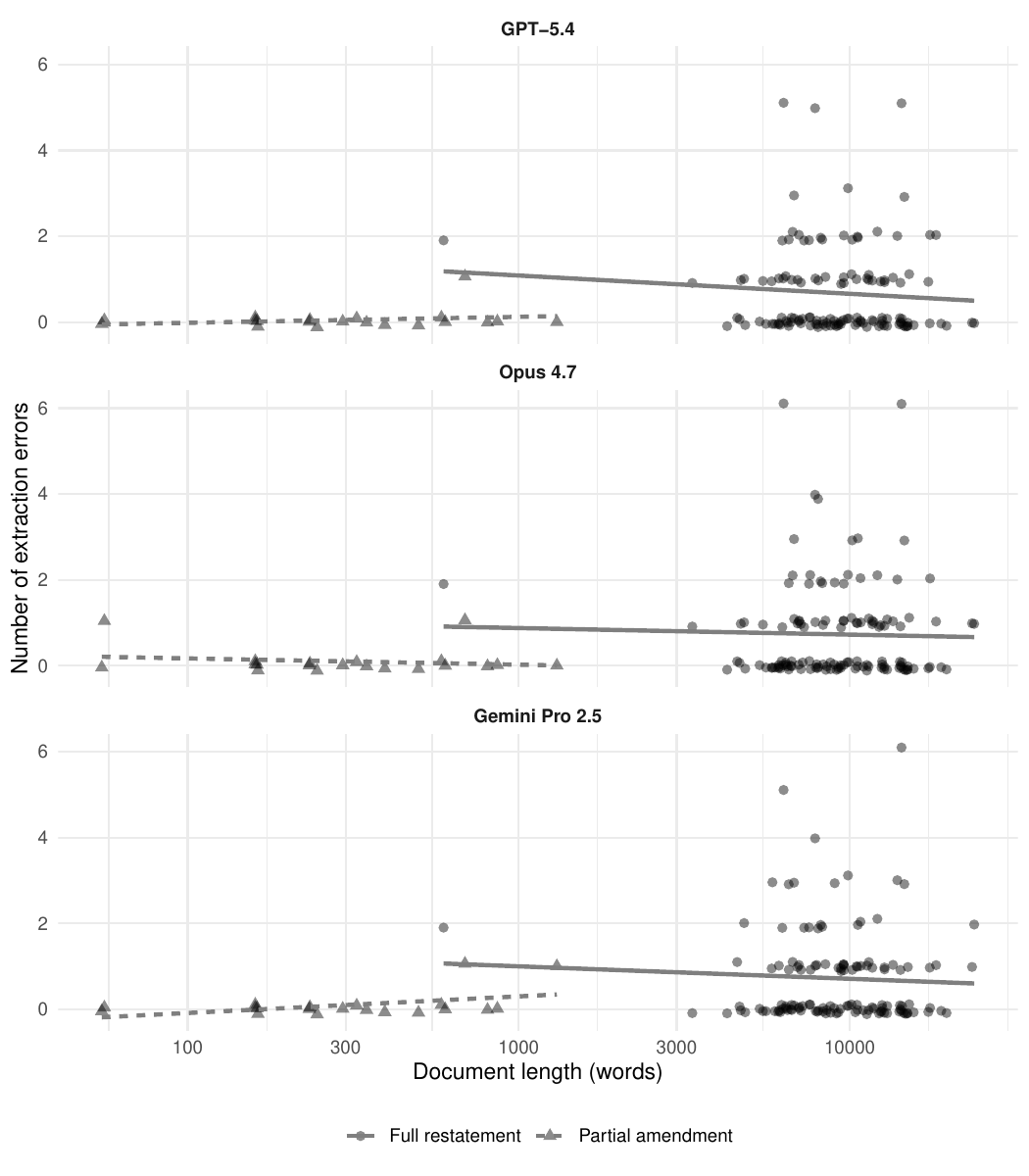}
\caption{Error distribution vs bylaw length.}
\label{fig:errordistribution2}
\end{figure*}

More importantly, there is little evidence that extraction performance deteriorates with document length among full restatements. Despite substantial variation in the length of these documents, the line representing a fitted linear relationship between document length and the number of extraction errors is flat to slightly negative for each of the three frontier models. Similarly, the documents with the highest error counts for all three models are roughly evenly distributed over the range of document length, with two documents in the lower half and one document in the upper half. Interestingly, the same three bylaws account for the highest error counts across all three models.

Formal overdispersion tests confirm that the distribution of errors among documents is not random, but that some documents appear to pose harder extraction problems. Figure \ref{fig:errordistribution1} illustrates this finding by comparing the observed distribution of error counts among full restatements with the binomial distribution implied by a constant error probability across documents. For all three frontier models, the observed error counts are substantially overdispersed relative to this benchmark, with a small number of bylaws accounting for much of the excess dispersion.  

\begin{figure*}[htb!]
\centering
\includegraphics[width=.7\textwidth]{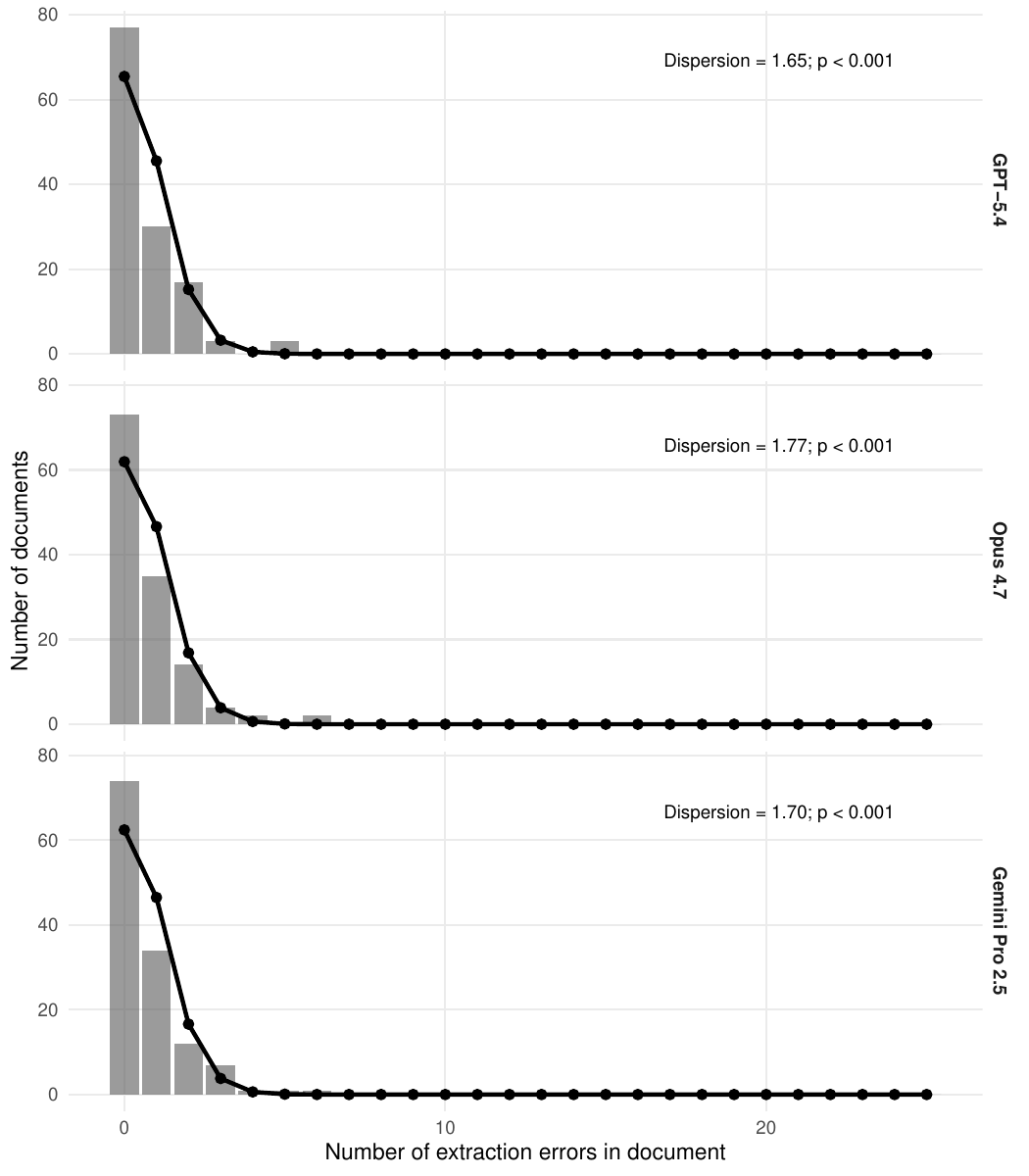}
\caption{Error distribution among full restatements.}
\label{fig:errordistribution1}
\end{figure*}

Together, these results provide additional evidence that, over the range of document lengths commonly found in corporate bylaws, providing the full document to the model does not appear to impair extraction performance.
\clearpage
\section{Extraction Prompts}
\label{app:pipelines}

This Appendix reports two prompts illustrating the differences between the simple and detailed prompts. For variable $SB$, the simple prompt contains only the following instructions:

\small

\begin{lstlisting}[breaklines=true]
Detailed Questions:
- Staggered Board:
  * currently_staggered: Does the bylaw stipulate that the corporation's board is staggered/classified?
\end{lstlisting}

\normalsize
By contrast, the detailed prompts include the following more detailed questions and instructions:

\small

\begin{lstlisting}[breaklines=true]
Detailed Questions for JSON Fields:
- Contains Staggered Board Provision: Do the bylaws include any provision addressing staggered boards or the division of directors into different classes? Answer "Y" or "N".
- Staggered Board:
  * Staggered Board: Do the bylaws expressly stipulate that the board is staggered or that directors are divided into classes? "Y"/"N".
Scope/Interpretation:
- Contains Staggered Board Provision = "Y" if the excerpt discusses director classes in any sense (including share-class seat allocation), even if Staggered Board ultimately = "N" under the staggered-board test.
- Staggered/classified board test: Code Staggered Board="Y" only if the bylaws state or clearly imply that directors are divided into classes with multi-year terms and that only a subset of directors is elected at each annual meeting (e.g., "one class each year").
- Exclude share-class seat allocation: If "classes" language only allocates director election rights or board seats by classes/series of stock (e.g., "Class A elects X directors" or "Series A Preferred elects Y directors"), do not treat this as a staggered board.
\end{lstlisting}

\normalsize
For variables in the $REM$ family, the simple prompts contain the following questions:

\small

\begin{lstlisting}[breaklines=true]
Detailed Questions:
- Director_Removal:
  * mentioned: Do the bylaws mention the removal of directors?
  * for_cause_requirement: Do the bylaws stipulate that directors can only be removed for cause?
  * general_supermajority: Do the bylaws stipulate that a supermajority of shareholder votes is required to remove a director without cause? If yes, please return the percentage requirement.
  * for_cause_supermajority: Do the bylaws stipulate that a supermajority of shareholder votes is required to remove a director with cause? If yes, please return the percentage requirement.
\end{lstlisting}

\normalsize
And these are the corresponding instructions in the detailed prompts:

\small
\begin{lstlisting}[breaklines=true]
Detailed Questions for JSON Fields:
- Contains Removal Provision: Do the bylaws include any provision addressing the removal of directors? Answer "Y" or "N".
- Removal:
  * For cause: Do the bylaws allow director removal for cause? "Y"/"N".
  * Supermajority Threshold (%) (removal with cause): Return the integer percent required only if the bylaw requires >50% to remove with cause; otherwise return "N". Return "NA" if For cause = "N".
  * Votes based on outstanding (removal with cause): "Y" if the removal vote is keyed to outstanding/entitled-to-vote shares; otherwise "N". If silent, "N". Return "NA" if For cause = "N".
  * For cause only: Do the bylaws state directors may be removed only for cause? "Y"/"N"/"NA".
  * For cause only conditional on classified board: If "for cause only" applies only when the board is classified/staggered, set "Y"; else "N"; "NA" unless For cause only="Y".
  * Without cause: Do the bylaws affirmatively allow removal without cause (e.g., "with or without cause")? "Y"/"N"/"NA" (apply gating above).
  * Supermajority Threshold (%) (removal without cause): If removal without cause is permitted, return the integer percent required only if the bylaw requires >50% to remove without cause; otherwise return "N". If removal without cause is not permitted/inapplicable, return "NA".
  * Votes based on outstanding (removal without cause): If removal without cause is permitted and the bylaw specifies the denominator, return "Y" for outstanding/entitled-to-vote; otherwise "N". Return "NA" if removal without cause is not permitted/inapplicable.
  * Text Quote: Minimal clause(s) supporting your answers ("" if all fields are "N"/"NA").
\end{lstlisting}

\clearpage
\endgroup

\end{document}